\documentclass{article}

\usepackage{microtype}
\usepackage{graphicx}
\usepackage{booktabs} 

\usepackage{hyperref}

\usepackage[arxiv]{icml2023}

\usepackage{amsmath}
\usepackage{amssymb}
\usepackage{mathtools}
\usepackage{amsthm}

\usepackage[capitalize,noabbrev]{cleveref}

\theoremstyle{plain}
\newtheorem{theorem}{Theorem}[section]

\newtheorem{lemma}[theorem]{Lemma}

\theoremstyle{definition}
\newtheorem{definition}[theorem]{Definition}
\newtheorem{assumption}[theorem]{Assumption}
\theoremstyle{remark}

\usepackage{amsmath,amsfonts,bm}

\def\eqref#1{equation~\ref{#1}}

\def\1{\bm{1}}

\DeclareMathAlphabet{\mathsfit}{\encodingdefault}{\sfdefault}{m}{sl}
\SetMathAlphabet{\mathsfit}{bold}{\encodingdefault}{\sfdefault}{bx}{n}

\usepackage[list=true]{subcaption}
\usepackage{hyperref}
\usepackage{url}
\usepackage{amsmath,amsthm}
\usepackage{xspace}
\usepackage{enumitem}
\usepackage{caption}
\usepackage{subcaption}
\usepackage{multirow}
\usepackage{titletoc}
\usepackage{graphicx}
\usepackage{booktabs} 
\usepackage{physics}
\usepackage{wrapfig,lipsum,booktabs}
\newcommand{\faa}{\mathcal{F\!A\!A}}

\usepackage{float}
\newcommand{\cifar}{CIFAR\xspace}
\newcommand{\mnist}{MNIST\xspace}
\newcommand{\name}{FOCUS\xspace}
\newcommand{\fairname}{FAA\xspace}

\definecolor{revcolor}{HTML}{000000}

\newcommand{\revise}[1]{\textcolor{revcolor}{#1}}

\usepackage[textsize=tiny]{todonotes}

\icmltitlerunning{\name: Fairness via Agent-Awareness for Federated Learning on Heterogeneous Data}

\begin{document}

\twocolumn[
\icmltitle{\name: Fairness via Agent-Awareness for Federated Learning on Heterogeneous Data}



\icmlsetsymbol{equal}{*}

\begin{icmlauthorlist}
\icmlauthor{Wenda Chu}{equal,tsinghua}
\icmlauthor{Chulin Xie}{equal,uiuc}
\icmlauthor{Boxin Wang}{uiuc}
\icmlauthor{Linyi Li}{uiuc}
\icmlauthor{Lang Yin}{uiuc}
\icmlauthor{Arash Nourian}{amazon}
\icmlauthor{Han Zhao}{uiuc}
\icmlauthor{Bo Li}{uiuc}
\end{icmlauthorlist}

\icmlaffiliation{tsinghua}{
California Institute of Technology}
\icmlaffiliation{uiuc}{University of Illinois Urbana-Champaign}
\icmlaffiliation{amazon}{Amazon}


\icmlkeywords{Machine Learning, ICML}

\vskip 0.3in
]



\printAffiliationsAndNotice{\icmlEqualContribution} 

\begin{abstract}
Federated learning (FL) allows agents to jointly train a global model without sharing their local data.
However, due to the heterogeneous nature of local data, existing definitions of fairness in the context of FL are prone to noisy agents in the network.
For instance, existing work usually considers accuracy parity as fairness for different agents in FL, which is not robust under the heterogeneous setting, since it will enforce agents with high-quality data 
to achieve similar accuracy to those who contribute low-quality data,
 which may discourage the benign agents from participating in FL.
In this work, we propose a formal FL fairness definition, \textit{fairness via agent-awareness} (\fairname), which takes the heterogeneity of different agents into account. Under \fairname, the performance of agents with high-quality data will not be sacrificed just due to the existence of large numbers of agents with low-quality data.
In addition, we propose a fair FL training algorithm based on agent clustering (\name) to achieve fairness in FL, as  measured by \fairname. Theoretically, we prove the convergence and optimality of \name under mild conditions for both linear and general convex loss functions with bounded smoothness. We also prove that \name always achieves higher fairness in terms of \fairname compared with standard FedAvg under both linear and general convex loss functions. Empirically, we show that on four FL datasets, including synthetic data, images, and texts, \name achieves significantly higher fairness in terms of \fairname while maintaining competitive prediction accuracy compared with FedAvg and state-of-the-art fair FL algorithms. 
\end{abstract}

\vspace{-4mm}
\section{Introduction}
\vspace{-1mm}
Federated learning (FL) is emerging as a promising approach to enable scalable intelligence over distributed settings  such as mobile networks~\citep{lim2020federated, hard2018federated}. 
Given the wide adoption of FL in medical analysis~\citep{Sheller2020Med, Adnan2022Med}, recommendation systems~\citep{Minto2021Rec, Anelli2021Rec}, and personal Internet of Things (IoT) devices~\citep{IOT-Healthcare}, it has become a central question on how to ensure the fairness of the trained global model in FL networks before its large-scale deployment by local agents, especially when the data quality/contributions of different agents are  different in the heterogeneous setting.  

In standard (centralized) ML, fairness is usually defined as a notion of parity of the underlying distributions from different groups given by a protected attribute (e.g., gender, race). Typical definitions include demographic parity~\citep{zemel2013learning,dwork2012fairness}, equalized odds~\citep{hardt2016equality}, and accuracy parity~\citep{buolamwini2018gender,zhao2019inherent}. However, it is yet unclear what is the desired notion of fairness in FL. Previous works that explore fairness in FL mainly focus on the demographic disparity of the final trained model regarding the protected attributes without considering different contributions of agents~\citep{FedFair, Hu2022ProvablyFair} or the accuracy disparity across agents~\citep{Li2019fair, Donahue2022, Mohri2019AFL}. 
Some works have taken into consideration the local data properties~\citep{Zhang2020Hierarchy, Kang2019Mobile} and data size~\citep{fair-fl-size}. However, the fairness analysis in FL under heterogeneous agent contributions is still lacking. Thus, in this paper, we aim to ask: \textit{What is a desirable notion of fairness in FL?} On one hand, it should be able to take the heterogeneity of different local agents into account so that it is robust to the potential noisy nodes in the network. On the other hand, it should also encourage the participation of benign agents that contribute high-quality data to the FL network. Furthermore, \emph{can we design efficient training algorithms to guarantee it?}

In light of the above questions, in this work, we aim to define and enhance fairness by explicitly considering the  heterogeneity of local agents. In particular,  
for FL trained with standard FedAvg protocol~\citep{fedavg}, if we denote the data of agent $e$ as $D_e$ with size $n_e$ and the total number of data as $n$, the final trained global model aims to minimize the loss with respect to the global distribution $\mathcal P=\sum_{e=1}^E \frac{n_e}{n} D_e$, where $E$ is the total number of agents. 
In practice, some local agents may have low-quality data (e.g., free riders), so intuitively it is ``unfair" to train the final model regarding such global distribution over all agents, which will sacrifice the performance of agents with high-quality data.
For example, considering the FL applications for medical analysis, some hospitals have high-resolution medical data with fine-grained labels, which are expensive to collect; some hospitals may only be able to provide low-resolution medical data and noisy labels. In such a setting, the utility of agents with high-quality data would be harmed due to the aggregated global distribution.
Thus, a proper fairness notion is critical to encourage agents to participate in FL and ensure their utility.

In this paper, we define \textbf{fairness via agent-awareness in FL (\fairname)} as $\faa(\{\theta_e\}_{e\in [E]}) = \max_{e_1,e_2\in E} \big|\mathcal E_{e_1}(\theta_{e_1}) - \mathcal E_{e_2}(\theta_{e_2}) \big|$, where $\mathcal{E}_{e}$ is the \textit{excess risk} of an agent $e \in E$ with model parameter $\theta_e$. Technically, 
the excess risk of each agent is calculated as $\mathcal E_e(\theta_e) = \mathcal L_e(\theta_e) - \min_{\theta^*}\mathcal L_e(\theta^*)$, which stands for the loss of user $e$ evaluated on the FL model  $\theta_e$
subtracted by the Bayes optimal error of the local data distribution ~\citep{opper1991generalization}. 
\revise{For each agent, a lower excess risk $\mathcal E_e(\theta_e)$ indicates a \emph{higher gain} from the FL model $\theta_e$ w.r.t the local distribution because its loss $\mathcal L_e(\theta_e)$ is closer to its Bayes optimal error.} 
\revise{Notably, reducing FAA enforces the \textit{equity of excess risks} among agents, following the classic philosophy that agents \textit{``do not suffer from scarcity, but inequality of gains''} (gains in terms of \textit{relative} performance improvement measured by excess risks) from participating in FL.} Therefore, lower \fairname indicates stronger fairness for FL. 

Based on our fairness definition \fairname, 
we then propose a \textit{fair FL algorithm based on agent clustering} (\name) to improve the fairness of FL. Specifically, we first cluster the local agents based on their data distributions and then train a model for each cluster. During inference time, the final prediction will be the weighted aggregation over the prediction result of each model trained with the corresponding clustered local data.  \underline{Theoretically}, we prove that the final converged stationary point of \name is exponentially close to the optimal cluster assignment under mild conditions. In addition, we prove that the fairness of \name in terms of \fairname is strictly higher than that of the standard FedAvg under both linear models and general convex losses.
\underline{Empirically}, we evaluate \name on four datasets, including  synthetic data, images, and texts, 
and we show that \name achieves higher fairness measured by \fairname than FedAvg and SOTA fair FL algorithms while maintaining similar or even higher prediction accuracy. 

\vspace{-1mm}
\underline{\textbf{Technical contributions}}. We define and improve FL fairness in heterogeneous settings by considering the different contributions of heterogeneous local agents. 
We make contributions on both the theoretical and empirical fronts. 
\vspace{-2mm}
\begin{itemize}[noitemsep,topsep=1pt,leftmargin=*]
    \item We formally define \textit{fairness via agent-awareness  (\fairname)} in FL based on agent-level excess risks to measure fairness in FL, and explicitly take the heterogeneity nature of local agents into account.
    \item We propose a fair FL algorithm via agent clustering (\name) to improve fairness measured by \fairname, especially in the heterogeneous setting. We prove the convergence rate and optimality of \name under linear models and general convex losses.
    \item Theoretically, we prove that \name achieves stronger fairness measured by \fairname compared with FedAvg for both linear models and general convex losses.
    \item Empirically, we compare \name with FedAvg and SOTA fair FL algorithms on four datasets, including synthetic data, images, and texts under heterogeneous settings. We show that \name indeed achieves stronger fairness measured by \fairname while maintaining similar or even higher prediction accuracy on all datasets.
\end{itemize}

\vspace{-3mm}
\section{Related work}
\vspace{-1mm}
\paragraph{Fair Federated Learning} There have been several studies exploring fairness in FL.
\citeauthor{Li2019fair} first define agent-level fairness by considering  \textit{accuracy equity} across agents and achieve fairness by assigning the agents with worse performance with higher aggregation weight during training. However, such a definition of fairness fails to capture the heterogeneous nature of local agents.
\citeauthor{Mohri2019AFL} 
pursue accuracy parity by improving the performance of the worst-performing agent. \citet{Wang2021FairAvg} propose to mitigate conflict gradients from local agents to enhance fairness. 
\revise{Instead of pursuing fairness with  one  single global model, \citeauthor{li2021ditto} propose to train a personalized model for each agent to achieve accuracy equity for the personalized models.}
\citeauthor{Zhang2020Hierarchy} predefine the agent contribution levels  based on
an oracle assumption (e.g., data volume,  data
collection cost, etc.) for fairness optimization, which lacks quantitative measurement metrics in practice.  
\citeauthor{xu2021gradient} approximate the Shapely Value based on gradient cosine similarity to evaluate agent contribution. However, \citeauthor{Zhang2020Hierarchy} point out that Shapely Value may discourage agents with rare data, especially under heterogeneous settings.
Here we provide an algorithm to quantitatively measure the contribution of local data based on each agent's excess risk, which will not be affected even if the agent is the minority.

\vspace{-4mm}
\paragraph{Clustered Federated Learning} 
Clustered FL algorithms are initially designed for multitasking and personalized federated learning, which assumes that agents can be naturally partitioned into clusters~\citep{ghosh2020efficient, multicenterxie, Sattler2021Cluster, marfoq2021federated}. 
Existing clustering algorithms usually aim to assign each agent to a cluster that provides the lowest loss~\citep{ghosh2020efficient}, optimize the clustering center to be close to the local model~\citep{multicenterxie}, or cluster agents with similar gradient updates (with respect to, e.g., cosine similarity~\citep{Sattler2021Cluster}) to the same cluster. In addition to these hard clustering approaches (i.e., each agent only belongs to one cluster), soft clustering has also been studied~\citep{marfoq2021federated, Li2021soft,Ruan2022FedSoft, Stallmann2022FFCM}, which enables the agents to benefit from multiple clusters. However, none of these works considers the fairness of clustered FL and the potential implications, and our work makes the first attempt to bridge them.
\section{Fair Federated Learning on Heterogeneous Data}
\label{sec:method}
\vspace{-2mm}
We first define our fairness via agent-awareness in FL with heterogeneous data and then introduce our fair FL based on the agent clustering (\name) algorithm to achieve \fairname.
\vspace{-2mm}
\subsection{Fairness via Agent-Awareness in FL (\fairname)}
\vspace{-2mm}
\label{subsec:3-1}
Given a set of $E$ agents participating in the FL network, each agent $e$ only has access to its local dataset: $D_e = \{(x_e,y_e)\}_{i=1}^{n_e}$, which is sampled from a distribution $\mathcal P_e$. The goal of standard FedAvg training is to minimize the overall loss $\mathcal L_E (\theta)$ based on the local loss $\mathcal L_e(\theta)$ of each agent:
\vspace{-5pt}
\begin{align}
\min_{\theta} \mathcal L_E(\theta) &= \sum\nolimits_{e\in [E]} \frac{|D_e|}{n} \mathcal L_e (\theta),\nonumber\\[-0.2em]
\mathcal L_e(\theta) &= \mathbb E_{(x,y)\in \mathcal P_e} \ell(h_{\theta}(x), y).
\end{align}
where $\ell(\cdot,\cdot)$ is a loss function given model prediction $h_{\theta}(x)$ and label $y$ (e.g., cross-entropy loss), $n =\sum\nolimits_{e\in [E]} |D_e| $ represents the total number of training samples, and $\theta$ represents the parameter of trained global model.

Intuitively, the performance of agents with high-quality and clean data could be severely compromised by the existence of large amounts of agents with low-quality and noisy data under FedAvg. 
To solve such a problem and characterize the distinctions of local data distributions (contributions) among agents to ensure fairness, we propose fairness via agent-awareness in FL (\fairname) as follows.
\begin{definition}[\bf Fairness via agent-awareness for FL (\fairname)]
\label{def:fairness}
Given a set of agents $E$ in FL, the overall fairness score among all agents is defined as the maximal difference of excess risks  for any pair of agents:

\vspace{-20pt}
\begin{equation}
   \faa(\{\theta_e\}_{e\in[E]}) =  \max_{e_1,e_2\in [E]} \Big|\mathcal E_{\revise{e_1}}(\theta_{e_1}) - \mathcal E_{e_2}(\theta_{e_2}) \Big|.
\end{equation}
\vskip -5pt

where $\theta_{e}$ is the local model for agent $e \in [E]$. The excess risk $\mathcal E_e(\theta_e)$ for agent $e$ given model $\theta_e$ is defined as the difference between the population loss $\mathcal L_e(\theta_e)$ and the Bayes optimal error of the corresponding data distribution, i.e.,
\begin{equation}
    \mathcal E_e(\theta_e) = \mathcal L_e(\theta_e) - \min_{\theta^*} \mathcal L_e(\theta^*),
\end{equation}
\vskip -10pt
where $\theta^*$ denotes any possible models.
\end{definition}

\vspace{-7pt}
Note that in FedAvg, each client uses the global model $\theta$ as its local model $\theta_e$.
\cref{def:fairness} represents a quantitative data-dependent measurement of agent-level fairness.
Instead of forcing accuracy parity among all agents regardless of their data quality, we define agent-level fairness as the equity of \textit{excess risks} among agents, which takes the contributions of local data into account by measuring their Bayes errors. 
For instance, when a local agent has low-quality data, although the corresponding utility loss would be high, the Bayes error of such low-quality data is also high, and thus the excess risk of the user is still low, enabling the agents with high-quality data to achieve low utility loss for fairness. 
According to the definition, we note that \textit{lower \fairname indicates stronger fairness among agents}.


\subsection{Fair Federated Learning on Heterogeneous Data via Clustering (\name)}
\label{subsec:3-2}

\paragraph{Method Overview.} 
To enhance the fairness of FL in terms of \fairname, we provide an agent clustering-based FL algorithm (\name) by partitioning agents conditioned on their data distributions.
Intuitively, grouping agents with similar local data distributions and similar contributions together helps to improve fairness, since it reduces the intra-cluster data heterogeneity. 
We will analyze the fairness achieved by \name and compare it with standard FedAvg both theoretically (\cref{subsec:4-2}) and empirically (\cref{sec:exp}).

Our \name algorithm (\cref{algorithm1}) leverages the Expectation-Maximization algorithm to perform agent clustering.  Define $M$ as the number of clusters and $E$  as the number of agents. The goal of \name is to simultaneously optimize the soft clustering labels $\Pi$ and model weights $W$. Specifically, $\Pi = \{\pi_{em}\}_{e\in [E], m\in [M]}$ are the dynamic soft clustering labels, representing the estimated probability that agent $e$ belongs to cluster $m$; 
$W = \{w_m\}_{m\in [M]}$ represent the model weights for $M$ data {clusters}.
Given $E$ agents with datasets $D_1,\dots, D_E$, our \name algorithm follows a two-step scheme that alternately optimizes $\Pi$ and $W$.

\vspace{-10pt}
\paragraph{E step.} 
Expectation steps update the cluster labels $\Pi$ given the current estimation of $(\Pi, W)$. At $k$-th communication round, the server broadcasts the $M$ {cluster} models to all agents. The agents calculate the expected training loss $\mathbb E_{(x,y)\in D_e} \ell(x,y;w_m^{\revise{(t)}})$ for each {cluster} model {$w_m^{\revise{(t)}}$,} $m\in [M]$, and then update the soft clustering labels $\Pi$ according to \cref{eq:Estep}.

\vspace{-10pt}
\paragraph{M step.}
The goal of M steps in \cref{eq:Mstep} is to minimize a weighted sum of empirical losses for all local agents. However, given distributed data, it is impossible to find its exact optimal solution in practice. Thus, we specify a concrete protocol in \cref{M:init} $\sim$ \cref{M:agg} to estimate the objective in \cref{eq:Mstep}.
{At $\revise{t}$-th communication round, for each cluster model $w_m^{\revise{(t)}}$ received from server,} each agent $e$ first initializes its local model $\theta_{em(0)}^{\revise{(t)}}$ as $w_m^{\revise{(t)}}$, and then updates the model using its own dataset. To reduce communication costs, each agent is allowed to run SGD locally for $\revise{K}$ {local steps} as shown in \cref{M:update}. After $\revise{K}$ {local steps}, each agent sends the updated models $\theta_{em\revise{(K)}}^{\revise{(t)}}$ back to the central server, and the server aggregates the models of all agents by a weighted average {based on the soft clustering labels $\{\pi_{em}\}$}. We provide theoretical analysis for the convergence and optimality of \name under these multiple local updates in \cref{sec:theoretical}.

\vspace{-15pt}
\small
\begin{align}
\text{Clients:}\quad &\theta_{em(0)}^{\revise{(t)}} = w_m^{\revise{(t)}}.\label{M:init}\\[-4pt]
&\theta_{em\revise{(k+1)}}^{\revise{(t)}} = \theta_{em\revise{(k)}}^{\revise{(t)}} - \eta_{\revise{k}} \nabla \sum_{i=1}^{n_e} \ell\Big(h_{ \theta_{em\revise{(k)}}^{\revise{(t)}}}(x_e^{(i)}), y_e^{(i)}\Big),\nonumber\\[-0.2em]
&\hspace{11em}\forall k=1,\dots, K-1. \label{M:update}\\[-4pt]
\text{Server:}\quad &w_m^{\revise{(t+1)}} = \sum_{e=1}^E \frac{\pi_{em}^{\revise{(t+1)}} \theta_{em\revise{(K)}}^{\revise{(t)}}}{\sum_{e^\prime=1}^E \pi_{e^\prime m}^{\revise{(t+1)}}}\label{M:agg}.
\end{align}
\normalsize

\vspace{-10pt}
\paragraph{Inference.} At inference time, each agent ensembles the $M$ models by a weighted average on their prediction probabilities, i.e., a agent $e$ predicts $\sum_{m=1}^M \pi_{em} h_{w_m}(x)$ for input $x$. Suppose a test dataset $D_e^{test}$ is sampled from distribution $\mathcal P_e$. The test loss can be calculated by

\vspace{-12pt}
\small
\begin{equation}
\mathcal L_{test}(W,\Pi) = \frac{1}{|D_e^{test}|} \sum_{(x,y)\in D_e^{test}} \ell\Big(\sum_{m=1}^{M} \pi_{em} h_w(x), y\Big) 
\end{equation}
\normalsize

\vspace{-8pt}
For unseen agents that do not participate in the training process, their clustering labels $\Pi$ are unknown. Therefore, an unseen agent $e$ computes its one-shot clustering label \small $\pi_{em}^{(1)}, m\in [M]$ \normalsize according to \cref{eq:Estep}, and outputs predictions \small $\sum_{m=1}^M \pi_{em}^{(1)} h_{w_m}(x)$ \normalsize for the test sample $x$.

\begin{algorithm}[t]
\begin{small}
\caption{EM clustered federated learning algorithm}
\label{algorithm1}
\begin{algorithmic}
\STATE {\bfseries Input:} Agents with data $\{D_i\}_{i\in [E]}$ and $M$ learning models.
\STATE Initialize weights $w_m^{(0)}$ and $\pi_{em}^{(0)} = \frac{1}{M}$ for $m\in [M]$ and $e\in [E]$.
\FOR{\revise{$t=0$ to $T-1$}}
\FOR{agent $e\in [E]$}
\FOR{model $m\in [M]$}
\STATE E step:
\vspace{-2em}
\begin{equation}
\label{eq:Estep}
     \pi_{em}^{\revise{(t+1)}}\leftarrow \frac{\pi_{em}^{\revise{(t)}} \exp(-\mathbb E_{(x,y)\in D_e} \ell(x,y;w_m^{\revise{(t)}}))}{\sum_{m=1}^M \pi_{em}^{\revise{(t)}} \exp(-\mathbb E_{(x,y)\in D_e}\ell(x,y;w_m^{\revise{(t)}}))}
\end{equation}
\vspace{-2em}
\ENDFOR
\ENDFOR
\FOR{model $m\in [M]$}
\STATE M step:  
\vspace{-1.5em}
\begin{equation}
\label{eq:Mstep}
    w_m^{\revise{(t+1)}} \leftarrow \arg\min_{w} \sum_{e=1}^E \pi_{em}^{\revise{(t+1)}} \sum_{i=1}^{n_e} \ell\Big(h_w(x_e^{(i)}), y_e^{(i)}\Big)
\end{equation}
\vspace{-1.5em}
\ENDFOR
\ENDFOR
\STATE \textbf{Return} model weights $w_m^{\revise{(T)}}$
\end{algorithmic}
\end{small}
\end{algorithm}
\vspace{-5mm}

\section{Theoretical Analysis of \name}
\vspace{-3pt}
\label{sec:theoretical}
In this section, we first present the convergence and optimality guarantees of our \name algorithm; and then prove that it improves the fairness of FL regarding \fairname. Our analysis considers linear models  and then extends to nonlinear models with smooth and strongly convex loss functions.

\subsection{Convergence Analysis}
\paragraph{Linear models.}
We first start with linear models to deliver the main idea of our analysis. Suppose there are $E$ agents, each with a local dataset \small$D_e = \{(x_e^{(i)},y_e^{(i)})\}_{i=1}^{n_e}, (e\in [E])$ \normalsize generated from a Gaussian distribution. Specifically, we assume each dataset $D_e$ has a mean vector $\mu_e\in \mathbb R^d$, and $(x_e^{(i)}, y_e^{(i)})$ is generated by \small $y_e^{(i)} = \mu_e^T x_e^{(i)} + \epsilon_e^{(i)}$\normalsize, where $x_e^{(i)}$ is a random vector \small $x_e^{(i)} \sim\mathcal N(0,\delta^2 I_d)$ \normalsize and the label $y_e^{(i)}$ is perturbed by some random noise \small $\epsilon_e^{(i)}\sim \mathcal N(0,\sigma^2)$\normalsize. Each agent is asked to minimize the mean squared error to estimate $\mu_e$, so the empirical loss function for a local agent given $D_e$ is
\begin{equation}
    \mathcal L_{emp}(D_e;w) = \frac{1}{n_e}\sum_{i=1}^{n_e} (w^T x_e^{(i)} - y_{e}^{(i)})^2.
\end{equation}
\normalsize

\vspace{-8pt}
We further make the following assumption about the heterogeneous agents.
\begin{assumption}[Separable distributions]
\label{linear-separable}
Suppose there are $M$ predefined vectors $\{w^*_i\}_{i=1}^M$, where for any $m_1,m_2\in [M]$, $m_1\neq m_2$, $\|w^*_{m_1} - w^*_{m_2}\|_2\geq R$. A set of agents $E$ satisfy separable distributions if they {can be} partitioned into $M$ subsets $S_1,\dots, S_M$ {such that,} for any agent $e\in S_m$, $\|\mu_e - w_m^*\|_2 \leq r < \frac{R}{2}$.
\end{assumption}

\cref{linear-separable} guarantees that the heterogeneous local data distributions are separable so that an optimal clustering {solution} exists, in which $\{w^*_1,\dots, w^*_M\}$ are the centers of clusters.  

We next present \cref{theorem:linear_converge} to demonstrate the linear convergence rate to the optimal cluster centers for \name. Detailed proofs can be found in \cref{appendixA1}.
\begin{theorem}
\label{theorem:linear_converge}
Assume the agent set E satisfies the separable distributions condition in Assumption \ref{linear-separable}.
Given trained $M$ models with $\pi_{em}^{(0)} = \frac{1}{M}$, $\forall e,m$. Under the natural initialization $w_m$ for each model $m\in [M]$, which satisfies $\exists \Delta_0 > 0, \|w_m^{(0)} - w_m^*\|_2 \leq \min_{m^\prime \neq m} \|w_{m}^{(0)} - w^*_{m^\prime}\|_2 - 2(r+ \Delta_0)$ and $|D_e| = O(d)$. If learning rate $\eta\leq \min(\frac{1}{4\delta^2},  \frac{\beta}{\sqrt{\revise{T}}})$,  \name converges by
\small
\vspace{-3pt}
\begin{align}
    &\pi_{em}^{\revise{(T)}} \geq \frac{1}{1+(M-1)\cdot\exp(-2R\delta^2 \Delta_0 \revise{T})}, \forall e\in S_m\label{eq:pi}\\
    &\mathbb E\|w_m^{\revise{(T)}} - w^*_m\|_2^2 \leq (1 - \frac{2\eta \gamma_m\delta^2}{M})^{KT} (\|w_m^{(0)} - w^*_m\|_2^2 + A)\nonumber\\[-0.2em]
    & \hspace{5em} +2MKr
    +  \frac{1}{2}M\delta^2 E\beta T^{-1/2} O(\revise{K}^3, \sigma^2).\label{eq:w}
\end{align}
\normalsize
where $\revise{T}$ is the total number of communication rounds; $\revise{K}$ is the number of local updates in each communication round; $\gamma_m = |S_m|$ is the number of agents in the $m$-th cluster, and

\vspace{-10pt}
\small
\begin{equation}
   A= \frac{2E\revise{K}(M-1)
   \delta^2}{(1 - \frac{2\eta \delta^2 \gamma_m}{M})^{\revise{K}} - \exp(-2R\delta^2 \Delta_0)}\tag{caused by initial inaccurate clustering}.
\end{equation}
\normalsize
\end{theorem}

\vspace{-3pt}
\begin{proof}[Proof sketch.]
    To prove this theorem, we first consider E steps and M steps separately to derive corresponding convergence lemmas (\cref{lemma:1,lemma:2}). In E steps, the soft cluster labels $\pi_{em}$ increase for all $e\in S_m$, as long as $\|w_m^{\revise{(t)}} - w^*_m\|_2 < \|w_{m^\prime}^{\revise{(t)}} - w^*_m\|_2, \forall m^\prime \neq m$. On the other hand, $\|w_m^{\revise{(t)}} - w^*_m\|$ is guaranteed to shrink linearly as long as $\pi_{em}$ is large enough for any $e\in S_m$. We then integrate \cref{lemma:1,lemma:2} and prove \cref{theorem:linear_converge} using an induction argument.
\end{proof}

\vspace{-5pt}
\textbf{Remarks.} \cref{theorem:linear_converge} shows the convergence of parameters $(\Pi, W)$ to a near-optimal solution. \cref{eq:pi} implies that the agents will be \textit{correctly clustered} since $\pi_{em}$ will converge to 1 as {the number of communication rounds} $K$ increases.
In \cref{eq:w}, the first term diminishes exponentially, while the second term $2M\revise{K}r$ reflects the intra-cluster distribution divergence $r$. The last term originates from the data heterogeneity among clients across different clusters. Its influence is amplified by the number of local updates ($O(\revise{K}^3)$) and will also diminish to zero as the number of communication rounds $\small \revise{T}$ goes to infinity.
Our convergence analysis is conditioned on the natural clustering initialization for model weights $\small w_m^{(0)}$ towards a corresponding cluster center $w^*_m$, which is standard in convergence analysis for a mixture of models~\citep{yan2017convergence,balakrishnan2017statistical}.

\vspace{-3pt}
\paragraph{Smooth and strongly convex loss functions.}
Next, we extend our analysis to a more general case of non-linear models with $L$-smooth and $\mu$-strongly convex loss function. 

\begin{assumption}[Smooth and strongly convex loss functions] 
\label{assum:smooth-convex}
The population loss functions $\mathcal L_{e}(\theta)$ for each agent $e$ is $L$-smooth, i.e., $ \|\nabla^2 \mathcal L_{e}(\theta)\|_2 \leq L$. The loss functions are $\mu$-strongly convex, if the eigenvalues $\lambda$ of the Hessian matrix $\nabla^2 \mathcal L_e(\theta)$ satisfy $
    \lambda_{\min}(\nabla^2 \mathcal L_{e}(\theta)) \geq \mu.$
\end{assumption}

We further make an assumption similar to \cref{linear-separable} for the general case:
\begin{assumption}[Separable distributions]
\label{assum:separable}
    A set of agents $E$ satisfy separable distributions if they can be partitioned into $M$ subsets $S_1,\dots, S_M$ with  $w_1^*, ..., w_M^*$ representing the center of each set respectively, and the optimal parameter $\theta^*$ of each local loss $\mathcal{L}_e$ (i.e., $\theta_e^* = \arg\min_\theta \mathcal L_{e}(\theta)$) satisfy
    $\|\theta_e^* - w_m^*\|_2\leq r$.
    In the meantime, agents from different subsets have different data distributions, such that
    $
        \|w_{m_1}^* - w_{m_2}^*\|_2\geq R, \forall m_1, m_2\in [M], m_1\neq m_2.
    $
\end{assumption}
    
    


\begin{theorem}
\label{theorem:convex_converge}
Assume the agent set E satisfies the separable distributions condition in Assumption \ref{assum:separable}. 
Suppose loss functions have bounded variance for gradients on local datasets, i.e., ${\footnotesize\mathbb E_{(x,y)\sim \mathcal D_e} [\|\nabla\ell (x,y;\theta) - \nabla \mathcal L_e(\theta)\|_2^2]\leq \sigma^2}$, and the population losses are bounded, i.e., $\mathcal L_e\leq G, \forall e\in [E]$.
With $\small \pi_{em}^{(0)} = \frac{1}{M}$, $\small \exists \Delta_0 > 0, \|w_m^{(0)} - w_m^*\|_2 \leq  \frac{\sqrt{\mu}R}{\sqrt{\mu}+\sqrt{L}} - r - \Delta_0$,
and the learning rate of each agent $\small \eta \leq \min(\frac{1}{2(\mu+L)}, \frac{\beta}{\sqrt{\revise{T}}})$, \name converges by
\small
\vspace{-3pt}
\begin{align}
    & \pi_{em}^{\revise{(T)}} \geq \frac{1}{1 + (M-1)\exp(-\mu R\Delta_0 \revise{T})},\ \forall  e\in S_m\label{eq:pi_evolve}\\
    & \mathbb E\|w_m^{\revise{(T)}} - w_m^*\|_2^2 \leq (1 - \eta A)^{KT} (\|w_m^{(0)} - w_m^*\|_2^2 + B) \nonumber\\[-.2em]
    &\hspace{7em} +O(Kr) + ME\beta O(\revise{K}^3, \frac{\sigma^2}{n_e})T^{-1/2}
    \label{eq:w_evolve}
\vspace{-3pt}
\end{align}
\normalsize
where $\revise{T}$ is the total number of communication rounds; $\revise{K}$ is the number of local updates in each communication round; $\gamma_m = |S_m|$ is the number of agents in the $m$-th cluster, and
\small
\vspace{-3pt}
\begin{equation}
    \underbrace{A = \frac{2\gamma_m}{M}\frac{\mu L}{\mu + L}}_{\text{related to convergence rate}}, 
    \underbrace{B = \frac{GMTE(\frac{4L}{\mu}+\frac{6}{\mu(\mu+L)})}{(1 - \eta A)^{\revise{K}} - \exp(-\mu R\Delta_0)}}_{\text{caused by the offset of initial clustering}}.
\end{equation}
\normalsize
\end{theorem}
\vspace{-10pt}
\begin{proof}[Proof sketch.]
    We analyze the evolution of parameters $(\Pi, W)$ for E steps in \cref{lemma:3} and M steps in \cref{lemma:4}. \cref{lemma:3} shows that the soft cluster labels $\pi_{em}$ increase for all $e\in S_m$ in E steps as long as $\|w_m - w_m^*\|_2 < \frac{\sqrt{\mu}R}{\sqrt{\mu} + \sqrt{L}} - r$; whereas \cref{lemma:4} guarantees that the model weights $w_m$ get closer to the optimal solution $w_m^*$ in M steps. We combine \cref{lemma:3,lemma:4} by induction to prove this theorem. Detailed proofs are deferred to \cref{appendixA2.2}.
\end{proof}
\vspace{-5pt}
\textbf{Remarks.} \cref{theorem:convex_converge} extends the convergence guarantee of  $(\Pi, W)$ from linear models (\cref{theorem:linear_converge}) to general models with smooth and convex loss functions. For any agent $e$ that belongs to a cluster $m$ ($e\in S_m$), its soft cluster label $\pi_{em}$ converges to 1 based on \cref{eq:pi_evolve}, indicating the clustering optimality. Meanwhile, the model weights $W$ converge linearly to a near-optimal solution. The error term $O(\revise{K}r)$ in \cref{eq:w_evolve} is expected since $r$ represents the data divergence within each cluster and $w_m^*$ denotes the center of each cluster. The last term in \cref{eq:w_evolve} implies a trade-off between communication cost and convergence speed. Increasing $\revise{K}$ reduces communication cost by $O(\frac{1}{\revise{K}})$ but at the expanse of slowing down the convergence.

\vspace{-2pt}
\subsection{Fairness Analysis}
\vspace{-2pt}
\label{subsec:4-2}
To theoretically show that \name achieves stronger fairness in FL based on \fairname, here we focus on a simple yet representative case where all agents share similar distributions except one outlier agent. 

\vspace{-5pt}
\paragraph{Linear models.}
We first concretize such a scenario for linear models.
Suppose we have $E$ agents learning weights for $M$ linear models. Their local data $D_e (e\in [E])$ are generated by \small $y_e^{(i)} = \mu_e^T x_e^{(i)} - \epsilon_e^{(i)}$ \normalsize with \small $x_e^{(i)}\sim \mathcal N(0,\delta^2 I_d)$ \normalsize and \small $\epsilon_e^{(i)}\sim \mathcal N(0,\sigma_e^2)$. \normalsize
$E-1$ agents learn from a normal dataset with ground truth vector $\mu_1,\dots, \mu_{E-1}$ and $\|\mu_e - \mu^*\|_2\leq r$, while the $E$-th agent has an outlier data distribution, with its the ground truth vector $\mu_E$ far away from other agents, i.e., $\|\mu_E - \mu^*\|_2\geq R$.

As  stated in \cref{theorem:linear_converge}, the {soft clustering labels and} model weights $(\Pi, W)$ converge linearly to the global optimum. Therefore, we analyze the fairness of \name, assuming an optimal $(\Pi, W)$ is reached.
We compare the \fairname achieved by \name and FedAvg to underscore how our algorithm helps improve fairness for heterogeneous  agents.

\begin{theorem}
\label{fairness_linear}
When a single agent has an outlier distribution, the fairness \fairname achieved by \name algorithm with two clusters $M=2$ is
\begin{equation}
    \faa_{focus}(W,\Pi) \leq \delta^2 r^2.
\end{equation}
while the fairness \fairname achieved by FedAvg  is
\begin{equation}
    \faa_{avg}(W) \geq \delta^2\Big(\frac{R^2 (E-2) - 2Rr}{E} + r^2 \Big) = \Omega(\delta^2 R^2).
\end{equation}
\end{theorem}

\vspace{-8pt}
\textbf{Remarks.}
When a single outlier exists, the fairness gap between  Fedavg  and \name is shown by \cref{fairness_linear}.

\vspace{-8pt}
\small
\begin{equation}
    \faa_{avg}(W) - \faa_{focus}(W, \Pi) \geq \delta^2 \Big(\frac{R^2 (E-2) - 2Rr}{E}\Big).
\end{equation}
\normalsize

\vspace{-8pt}
As long as $R > \frac{2r}{E-2}$, \name is guaranteed to achieve stronger fairness (i.e., lower \fairname) than FedAvg. Note that \textit{the outlier assumption only makes sense when $E > 2$} since one cannot tell which agent is the outlier when $E = 2$. Also, we naturally assume $R > 2r$ so that the two underlying clusters are at least separable. Therefore, we conclude that \name dominates than FedAvg in terms of \fairname. Here we only discuss the scenario of a single outlier agent for clarity, but similar conclusions can be drawn for multiple underlying clusters and $M>2$, as discussed in \cref{appendix:B1}.

\vspace{-6pt}
\paragraph{Smooth and strongly convex loss functions.}
We generalize the fairness analysis to nonlinear models with smooth and convex  loss functions. To illustrate the superiority of our \name algorithms in terms of \fairname fairness, we similarly consider training in the presence of an outlier agent.
Suppose we have $E$ agents that learn weights for $M$ models. We assume their population loss functions are $L$-smooth, $\mu$-strongly convex (as in \cref{assum:smooth-convex}) and bounded, i.e., $\mathcal L_{e}(\theta)\leq G$.
$E-1$ agents learn from similar data distributions, such that the total variation distance between the distributions of any two different agents $i,j \in [E-1]$ is no greater than $r$: $D_{TV}(\mathcal P_i,\mathcal P_j) \leq r$. On the other hand, the $E$-th agent has an outlier data distribution, such that the Bayes error $\mathcal L_{E}(\theta_i^*) - \mathcal L_{E}(\theta_E^*) \geq R$ for any $i\in [E-1]$. We claim that this assumption can be reduced to a lower bound on H-divergence \citep{zhao2022comparing} between distributions $\mathcal P_i$ and $\mathcal P_E$ that $D_H(\mathcal P_i,\mathcal P_E)\geq \frac{LR}{4\mu}$. (See proofs in \cref{appendix:B3}.)

\begin{theorem}
\label{fairness_convex}
The fairness \fairname achieved by \name with two clusters $M=2$ is
\begin{equation}
    \faa_{focus}(W,\Pi) \leq \frac{2Gr}{E-1}
\end{equation}

\vspace{-12pt}
Let $B =\frac{2Gr}{E-1}$. The fairness achieved by FedAvg is
\vspace{-14pt}

\small
\begin{align}
\label{fedavg_fairness}
    \faa_{avg}(W)&\geq \Big(\frac{E-1}{E} - \frac{L}{\mu E^2}\Big) R
    - \Big(1 + \frac{L(E-1)}{\mu E} - \frac{L^2}{\mu^2 E}\Big)  B
    \nonumber\\
    &\hspace{8em}-\frac{2L}{\mu E}\sqrt{B(R - \frac{L}{\mu} B)}
\end{align}
\end{theorem}

\paragraph{Remarks.} 
Notably, when the outlier distribution is very different from the normal distribution, such that $R \gg Gr$ (which means $B \ll R$), we simplify \cref{fedavg_fairness} as
\begin{equation*}
 \faa_{avg}(W) \geq \left(\frac{E-1}{E} - \frac{L}{\mu E^2}\right)R.   
\end{equation*}
Note that $\faa_{focus}(W,\Pi) \leq B \ll R$, so the fairness \fairname achieved by FedAvg  is always larger (weaker) than that of \name, as long as $E \geq \sqrt{L/\mu}$, indicating the effectiveness of \name.

\vspace{-2mm}
\section{Experimental Evaluation}
\vspace{-1mm}
\label{sec:exp}
We conduct extensive experiments on various heterogeneous data settings to evaluate the fairness measured by \fairname for \name, FedAvg~\citep{fedavg}, and two baseline fair FL algorithms (i.e., q-FFL~\citep{Li2019fair} and AFL~\citep{Mohri2019AFL}). We show that \name achieves significantly higher fairness measured by \fairname while maintaining similar or even higher accuracy.
\vspace{-3pt}

\begin{table*}[ht]
    \centering
    \caption{ Comparison of \name, FedAvg, and  fair FL algorithms q-FFL, AFL, \revise{Ditto and CGSV}, in terms of average test accuracy (Avg Acc), average test loss (Avg Loss),  fairness \fairname  \revise{and existing fairness metric Agnostic loss}. \name achieves the best fairness measured by \fairname compared with all baselines.
    }
    \vspace{-5pt}
    \label{table:compare}
    \resizebox{.8\linewidth}{!}{
    \begin{tabular}{llcccccccc}
    \toprule
      && \multirow{2}{*}{FOCUS} & \multirow{2}{*}{FedAvg} & \multicolumn{3}{c}{q-FFL} & \multirow{2}{*}{AFL}  &\revise{\multirow{2}{*}{Ditto}} &\revise{\multirow{2}{*}{CGSV}}\\\cmidrule(lr){5-7}
     && & & $q=0.1$ & $q=1$ & $q=10$ & 
     \\
    \midrule
    \multirow{2}{*}[-2pt]{Synthetic} & Avg Loss & \bf 0.010 & 0.108 & 0.106 & 0.102 & 0.110 & 0.104 & \revise{0.023} & \revise{0.260}\\
     & FAA & \bf 0.001 & 0.958 & 0.769 & 0.717 & 0.699 & 0.780 & \revise{0.012} & \revise{0.010}\\
     \midrule
    \multirow{3}{*}[-2pt]{Rotated MNIST} & Avg Acc & \bf 0.953 & 0.929 & 0.922 & 0.861 & 0.685 & 0.885 & \revise{0.940} & \revise{0.938}\\
     & Avg Loss & \bf 0.152 & 0.246 & 0.269 & 0.489 & 1.084 & 0.429 & \revise{0.210} & \revise{0.222}\\
     & FAA & \bf 0.094 & 0.363 & 0.388 & 0.612 & 0.253 & 0.220 &\revise{0.104} & \revise{0.210}\\
     & \revise{Agnostic Loss} & \bf \revise{0.224} & \revise{0.616} & \revise{0.656} &\revise{1.018} &\revise{1.271} &\revise{0.548} &\revise{0.354} &\revise{0.331}\\
     \midrule
    \multirow{3}{*}[-2pt]{Rotated CIFAR}  & Avg Acc &  \bf 0.688 & 0.654 & 0.648 & 0.592 & 0.121 & 0.661 & 0.657 & 0.515\\
     & Avg Loss &  \bf 1.133  & 2.386 & 1.138 & 1.141 & 2.526 & 1.666 & 2.382 & 3.841\\
     & FAA & \bf 0.360 & 1.115 & 0.620 & 0.473 & 0.379 & 0.595 & 0.758 & 1.317\\
     & \revise{Agnostic Loss} & \bf\revise{1.294} & \revise{3.275} & \revise{1.610} & 1.439 & 2.526 & 2.179 & 3.053 & 3.841\\
     \midrule
    \multirow{3}{*}[-2pt]{Yelp/IMDb} & Avg Acc & {\bf 0.940} & {\bf 0.940} &  0.938 &   {0.938} &  0.909  &  0.934  & \revise{0.933} & \revise{0.701} \\
     & Avg Loss & \bf  0.174 & 0.236 & 0.188 & 0.179 &  0.264 &  0.187 & \revise{0.191} & \revise{0.547}  \\
     & FAA & \bf  0.047 & 0.098 & 0.052 & 0.051 &  0.070 & 0.049 & \revise{0.049} & \revise{0.462} \\
      & \revise{Agnostic Loss} & \revise{0.257} & \revise{0.349} & \revise{0.266} & \revise{0.253} & \bf \revise{0.242} & \revise{0.253} &  \revise{0.263} & \revise{0.700} \\
     \bottomrule
    \end{tabular}
    }
    \vspace{-10pt}
\end{table*}

\vspace{-5pt}
\subsection{Experimental Setup}
\label{subsec:setup}
\vspace{-1mm}
\textbf{Data and Models.}
We carry out experiments on four different datasets with heterogeneous data settings, ranging from synthetic data for linear models to images (rotated MNIST \citep{deng2012mnist} and rotated CIFAR \citep{cifar}) to text data for sentiment classification on Yelp \citep{yelp} and IMDb \citep{imdb} datasets. We train a fully connected model consisting of two linear layers with ReLU activations for \mnist, a ResNet 18 model \citep{He_2016_CVPR} for \cifar, and a pre-trained BERT-base model \citep{bert} for the text data. 
We refer the readers to Appendix~\ref{app:exp_details} for more implementation details.

\textbf{Evaluation Metrics and Implementation Details.}  We consider three evaluation metrics: average test accuracy, average test loss, \fairname for fairness, and  \revise{the existing fairness metric ``agnostic loss'' introduced by~\cite{Mohri2019AFL}}. 
For FedAvg, we evaluate the trained global model on each agent's test data; for \name, we train $M$ models corresponding to $M$ clusters, and use the soft clustering labels \small $\Pi = \{\pi_{em}\}_{e\in [E], m\in [M]}$ \normalsize to make aggregated predictions on each agent's test data.
We also report the performance of existing fair FL algorithms (i.e., q-FFL~\citep{Li2019fair},  AFL~\citep{Mohri2019AFL}, \revise{Ditto~\citep{li2021ditto}, and CGSV~\citep{xu2021gradient}}) as well as existing state-of-the-art FL algorithms in heterogeneous data settings (i.e., FedMA \citep{wang2020federated}, Bayesian nonparametric FL \citep{yurochkin2019bayesian} and FedProx \citep{li2020federated} in \cref{append:fairfl,append:SOTA}). 
 
To evaluate \fairname of different algorithms, we estimate the Bayes optimal loss $\min_w \mathcal L_e(w)$ for each local agent $e$. Specifically, we train a centralized model based on the subset of agents with similar data distributions (i.e., the same ground-truth cluster) and use it as a \textit{surrogate} to approximate the Bayes optimum.
We select the agent pair with the maximal difference of excess risks to measure fairness in terms of  \fairname calculated following \cref{def:fairness}.

\vspace{-2mm}
\subsection{Evaluation Results}
\vspace{-1mm}

\textbf{Synthetic data for linear models.}
We first evaluate \name on linear regression models with synthetic datasets. We fix $E = 10$ agents with data sampled from Gaussian distributions. Each agent $e$ is assigned with a local dataset of \small $D_e = \{(x_e^{(i)},y_e^{(i)})\}_ {i=1}^{n_e}$ \normalsize generated by $y_e^{(i)} = \mu_e^T x_e^{(i)} + \epsilon_e^{(i)}$ with \small $x_e^{(i)}\sim \mathcal N(0,I_d)$ \normalsize and \small $\epsilon_e^{(i)} \sim \mathcal N(0,\sigma^2)$\normalsize. We study the case considered in \cref{subsec:4-2} where a single agent has an outlier data distribution. We set the intra-cluster distance $r = 0.01$ and the inter-cluster distance $R = 1$ in our experiment. Note that it is a regression task, so we mainly report the average test loss instead of accuracy here. \cref{table:compare} shows that \name achieves \fairname of 0.001, much lower than the 0.958 achieved by FedAvg, 0.699 by q-FFL, and 0.780 by AFL.

\textbf{Rotated MNIST and CIFAR.}
Following~\citep{ghosh2020efficient},  we rotate the images \mnist and \cifar datasets with different degrees to create data heterogeneity among agents.
Both datasets are evenly split into 10 subsets for 10 agents. 
For \mnist, two subsets are rotated for 90 degrees, one subset is rotated for 180 degrees, and the rest seven subsets are unchanged, yielding an FL setup with three ground-truth clusters.  
Similarly, for \cifar, we fix the images of 7 subsets and rotate the other 3 subsets for 180 degrees, thus creating two ground-truth clusters. From \cref{table:compare}, we observe that \name consistently achieves higher average test accuracy, lower average test loss, and lower \fairname than other methods on both datasets. 
In addition, although existing fair algorithms q-FFL and AFL achieve lower FAA scores than FedAvg,  their average test accuracy drops significantly. This is mainly because these fair algorithms are designed for performance parity via improving low-quality agents (i.e., agents with high training loss), thus sacrificing the accuracy of high-quality agents. In contrast, \name improves both the \fairname fairness and preserves high test accuracy. 
\begin{figure*}[t]
     \centering
       \vspace{-2mm}
\subcaptionbox{{\footnotesize MNIST}}{\includegraphics[width=0.42\textwidth]{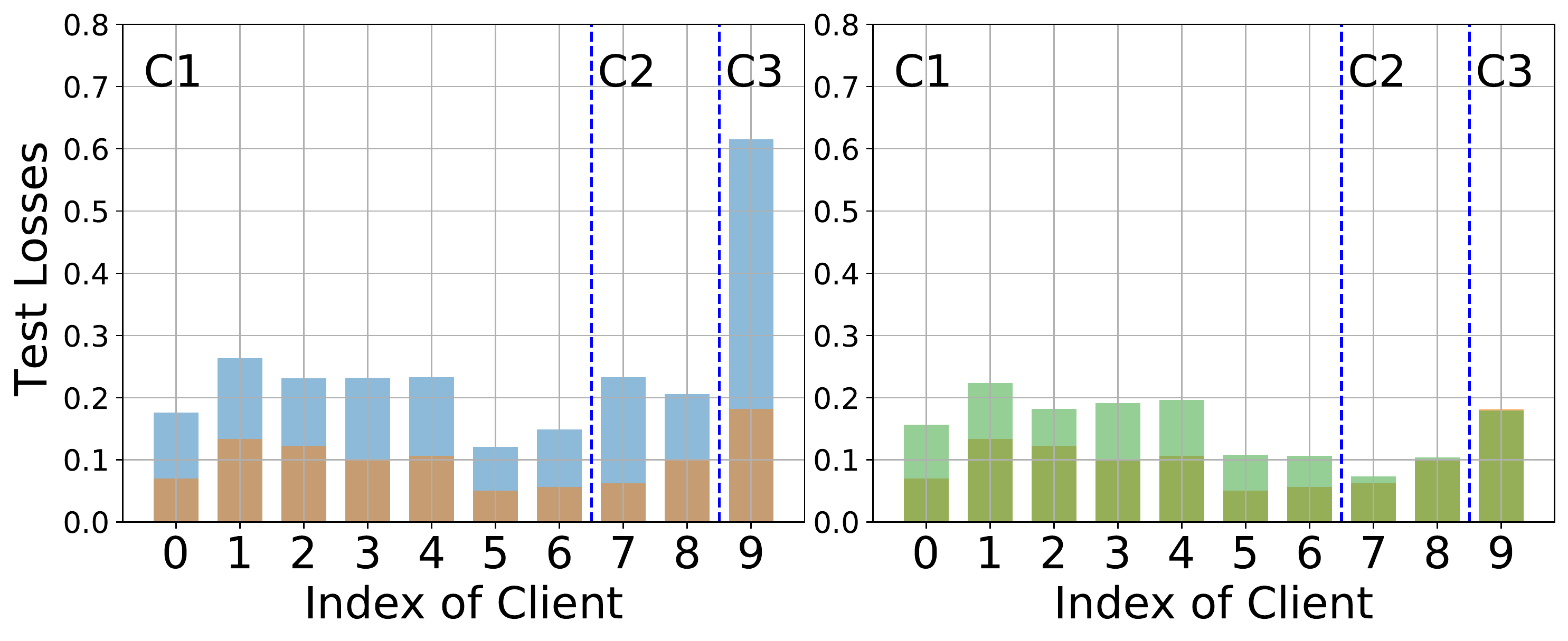}}%
\hfill
\subcaptionbox{{\footnotesize Yelp/IMDb}}{\includegraphics[width=0.55\textwidth]{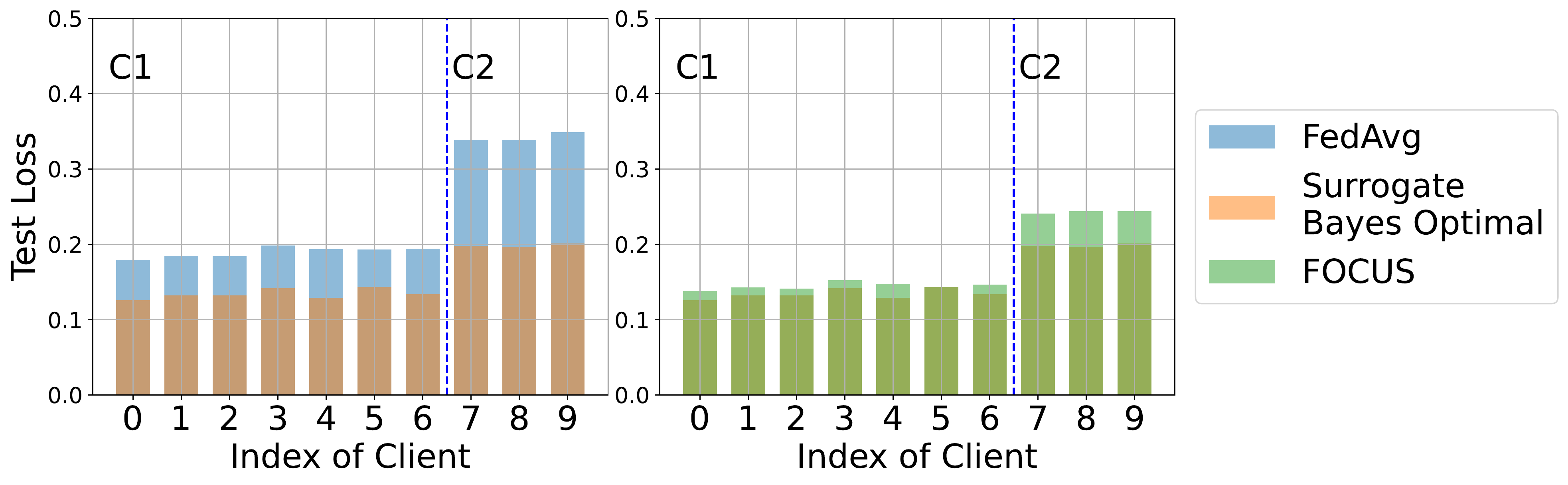}}%
\vspace{-9pt}
\caption{The excess risks of different agents trained with FedAvg and \name on MNIST (a) and Yelp/IMDb text data (b). $C_i$ denotes $i$th cluster.}
\label{fig:excess_risk}
\vspace{-13pt}
\end{figure*}

Next, we analyze the surrogate excess risk of every agent on \mnist in \cref{fig:excess_risk} (a).
We observe that the global model trained by FedAvg obtains the highest test loss as 0.61 on the outlier cluster, which rotates 180 degrees (i.e., cluster C3), resulting in high excess risk for the 9th agent. Moreover, the low-quality data of the outlier cluster affect the agents in the 1st cluster via FedAvg training, which leads to a much higher excess risk than that of \name. 
On the other hand,  \name successfully identifies clusters of the outlier distributions, i.e., clusters 2 and 3, rendering models trained from the outlier clusters independent from the normal cluster 1. 
As shown in \cref{fig:excess_risk}, our \name reduces the excess risks of all agents, especially the outliers, on different datasets. This leads to strong fairness among agents in terms of \fairname.  Similar trends are also observed in \cifar, in which our \name reduces the surrogate excess risk for the 9th agent from 2.74 to 0.44.  We omit  the loss histogram of \cifar  to \cref{append:CIFAR-loss}.

\begin{table}[ht]
\vspace*{-5pt}
\caption{Comparison of FOCUS and FedAvg with different numbers of outlier agents ($k$) in terms of average test accuracy (Avg Acc) and fairness \fairname.} 
\vspace{-7pt}
\label{table:num_outliers}
    \centering
    \small
    \scalebox{0.8}{
    \begin{tabular}{llcccccc}
    \toprule
        & & \multicolumn{3}{c}{Rotated MNIST} & \multicolumn{3}{c}{Rotated CIFAR} \\ \cmidrule(lr){3-5} \cmidrule(lr){6-8}
        & & $k=1$ & $k=3$ & $k=5$ & $k=1$ & $k=3$ & $k=5$\\
        \midrule
        \multirow{2}{*}{Avg Acc} & FOCUS & \textbf{0.957} & \textbf{0.953} & \textbf{0.948} & \textbf{0.683} & \textbf{0.688} & \textbf{0.677}\\
        & FedAvg & 0.945 & 0.929 & 0.910 & \textbf{0.683} & 0.654 & 0.651\\     \midrule
        \multirow{2}{*}{FAA} & FOCUS & \textbf{0.159} & \textbf{0.094} & \textbf{0.153} & \textbf{1.168} & \textbf{0.360} & \textbf{0.436}\\
        & FedAvg & 0.515 & 0.363 & 0.476 & 2.464 & 1.115 & 1.166\\
    \bottomrule
    \end{tabular}}
    \vspace{-10pt}
\end{table}

Additionally, we evaluate different numbers of outliers in \cref{table:num_outliers}. In the presence of 1, 3, and 5 outlier agents, forming 2, 3, or 4 underlying true clusters, FOCUS consistently achieves a lower FAA score and higher accuracy.  

In practice, we do not know the number of underlying clusters, so we set $M=2,3,4$ while we have 3 true underlying clusters in \cref{table:different_m} on \mnist. It shows that when $M=2,3,4$, \name achieves similar accuracy and fairness. When $M=1$, \name reduces to FedAvg, leading to the worst accuracy and fairness under heterogeneous data. A similar trend also occurs for \cifar, as shown in \cref{table:different_m}.

\begin{table}[ht]
    \centering
    \vspace{-3pt}
    \caption{The effect of the predefined number $M$ on Rotate \mnist with 3 underlying clusters and on Rotated \cifar with 2 underlying clusters.}
    \vspace{-7pt}
    \scalebox{0.65}{
    \begin{tabular}{lcccccccc}
    \toprule
         &  \multicolumn{4}{c}{Rotated MNIST} & \multicolumn{4}{c}{Rotated CIFAR}\\
         \cmidrule(lr){2-5} \cmidrule(lr){6-9}
         & \small $M=1$ & \small $M=2$ & \small $M=3$ & \small $M=4$ & \small  $M=1$ & \small $M=2$ &\small  $M=3$ & \small $M=4$\\
         \midrule
         Avg Acc & 0.929 & 0.952 & \bf 0.953 & \bf 0.953 & 0.654 & 0.688 & \bf 0.696 & 0.693\\
         Avg loss & 0.246 & 0.167 & \bf 0.152 & 0.153 & 2.386 & 1.133 & 0.932 & \bf 0.921\\
         FAA & 0.363 & \bf 0.079 & 0.094 & 0.091 & 1.115 & 0.360 & \bf 0.323 & 0.350 \\
         Agnostic & 0.616 & 0.272 & 0.224 & \bf 0.223 & 3.275 & 1.294 & 1.115 & \bf 1.098\\
         \bottomrule
    \end{tabular}}
    \label{table:different_m}
\end{table}

\vspace{-8pt}
\textbf{Sentiment classification.}
We evaluate \name on the sentiment classification task with text data,  Yelp (restaurant reviews), and IMDb (movie reviews), which naturally form data heterogeneity among $10$ agents and thus create $2$ clusters.
Specifically, we sample 56k reviews from Yelp datasets distributed among seven agents and use the whole 25k IMDB datasets distributed among three agents to simulate the heterogeneous setting. 
From \cref{table:compare}, we can see that while the average test accuracy of \name, FedAvg, and other fair FL algorithms are similar, \name achieves a lower average test loss. In addition, the \fairname of \name is significantly lower than other baselines, indicating stronger fairness. We also observe from \cref{fig:excess_risk} (b) that the excess risk of \name on the outlier cluster (i.e., C2) drops significantly compared with that of FedAvg. 

\begin{figure}[ht]
        \centering
        \vspace{-3pt}
        \includegraphics[width=0.475\linewidth]{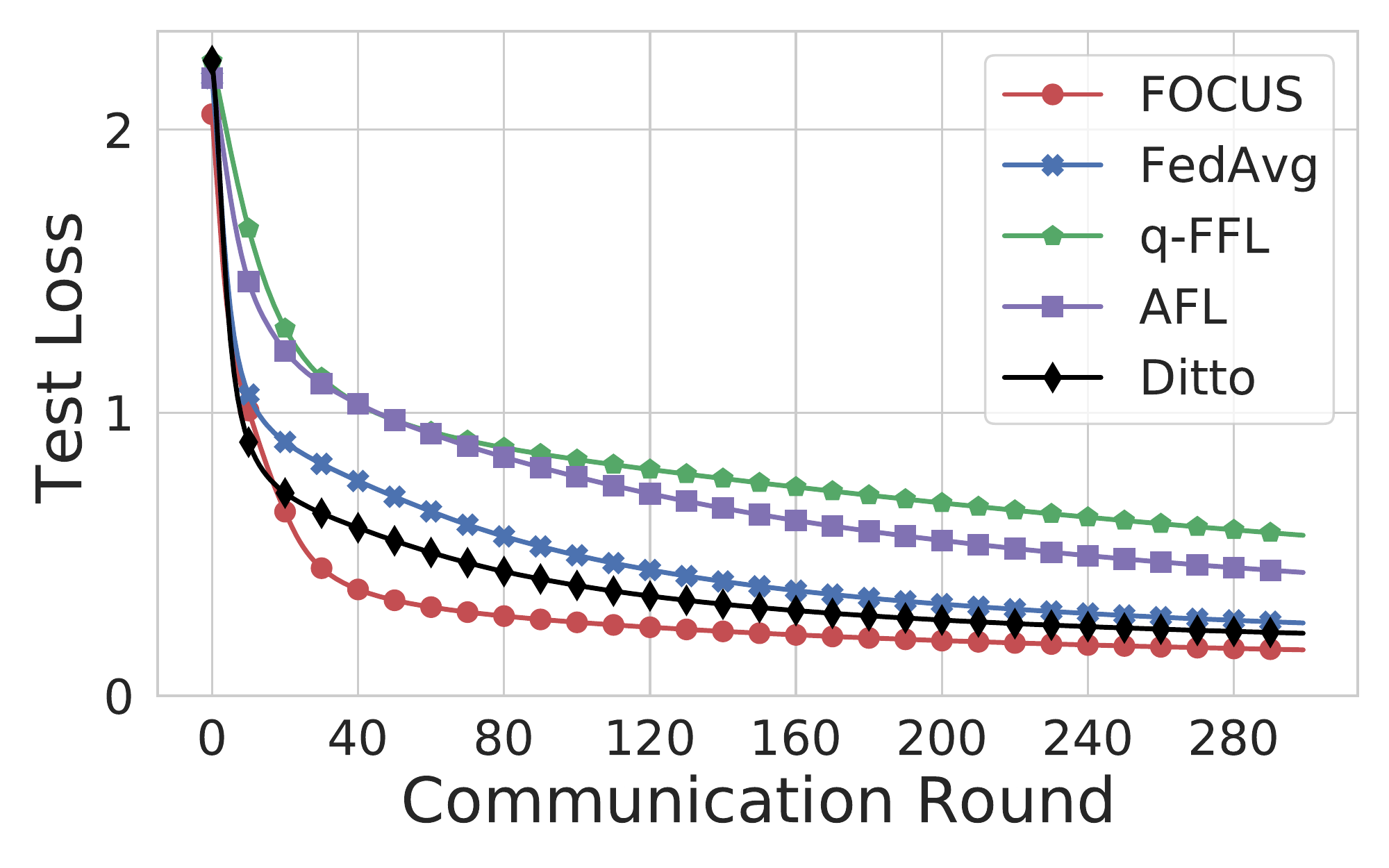}
        \hfill
        \includegraphics[width=0.475\linewidth]{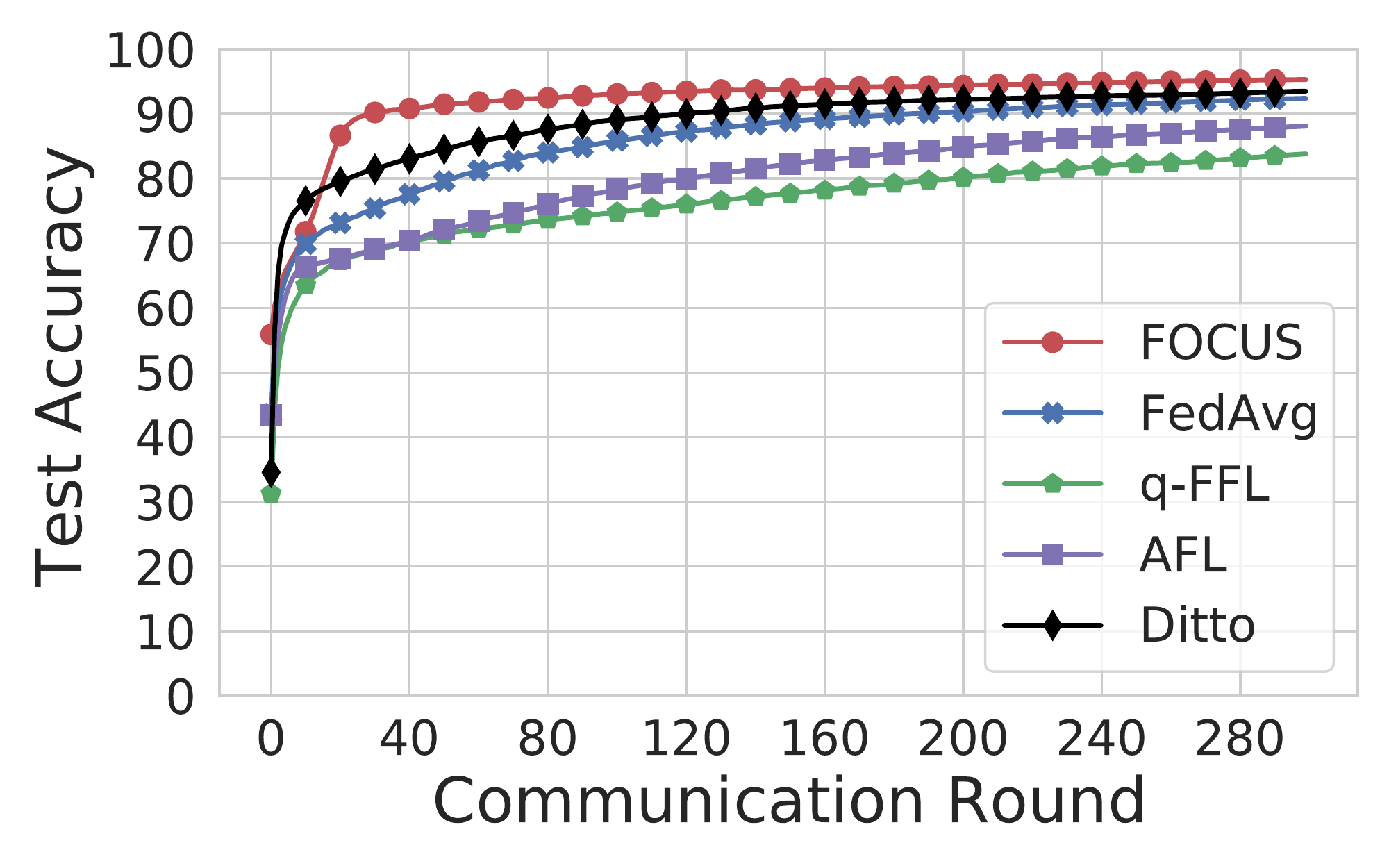}
        \vspace{-7pt}
    \caption{Comparison of the convergence rate of different methods on rotated \mnist with 100 clients.}
    \label{fig:cifar_converge}
\end{figure}

\textbf{Ablation Studies.} To provide a more comprehensive evaluation for \name, we present additional ablation studies on scalability, convergence rate, and runtime analysis. \cref{fig:cifar_converge} is a comparison of the convergence rate of different methods on rotated \mnist, which shows that \name converges faster and achieves better performance on MNIST under a large number of clients. Additional results in \cref{append:scale,append:convergence,append:runtime} show that \name is scalable to larger client groups and consumes comparable running time with other methods on different datasets. Moreover, we compare \name to a cluster-wise FedAvg algorithm with hard clustering, illustrating the advantages of \name using soft-clustering when the underlying clusters are not perfectly separable. We refer readers to \cref{append:hard_clustering} for further discussions.

\vspace{-8pt}

\vspace{-2mm}
\section{Conclusion}
\vspace{-2mm}
In this work, we provide an agent-level fairness measurement in FL (\fairname) by taking agents' inherent heterogeneous data properties into account. Motivated by our fairness definition in FL, we also provide an effective FL training algorithm \name to achieve high fairness. We theoretically analyze the convergence rate and optimality of \name, and we prove that under mild conditions, \name is always fairer than the standard FedAvg protocol. We conduct thorough experiments on synthetic data with linear models as well as image and text datasets on deep neural networks. We show that \name achieves stronger fairness than FedAvg and achieves similar or higher prediction accuracy across all datasets. We believe our work will inspire new research efforts on exploring the suitable fairness measurements for FL under different requirements.

\newpage

\bibliography{iclr2023_conference}
\bibliographystyle{icml2023}
\newpage
\appendix
\onecolumn

\section{Additional Experimental Results}
\label{sec_revision}

\subsection{Experimental Setups}
\label{app:exp_details}
Here we elaborate more details of our experiments.

\paragraph{Machines.} We simulate the federated learning setup on a Linux machine with AMD Ryzen Threadripper 3990X 64-Core CPUs and 4 NVIDIA GeForce RTX 3090 GPUs. 

\paragraph{Hyperparameters.}
For each FL experiment, we implement both \name algorithm and FedAvg algorithm using SGD optimizer with the same hyperparameter setting. 
Detailed hyperparameter specifications are listed in \cref{tab:hyper} for different datasets, including learning rate, the number of local training steps, batch size, the number of training epochs, etc.

\begin{table}[ht]
\centering
\caption{Dataset description and hyperparameters.}
\label{tab:hyper}
\resizebox{\columnwidth}{!}{%
    \begin{tabular}{cccccccccccccccc}
    \toprule
     Dataset  &  $\#$ training samples  &  $\#$ test samples &  $E$ &  $M$ &  batch size   & learning rate & local training epochs & epochs \\\midrule
    \mnist  & 60000 & 10000 & 10 & 3 & 6000 & 0.1 & 10 & 300 \\ \midrule
    \cifar & 50000 & 10000 & 10 & 2 & 100 & 0.1 & 2 & 300 \\ \midrule
     Yelp/IMDB  & 56000/25000 & 38000/25000 & 10 & 2 & 512 & 5e-5 & 2 & 3 \\
    \bottomrule
\end{tabular}%
}
\end{table}

\subsection{Comparison with existing fair FL methods}
\label{append:fairfl}
We present the full results of existing fair federated learning algorithms on our data settings in terms of FAA. 
The results in \cref{table:MNIST-ffl,table:CIFAR-ffl} show that FOCUS achieves the lowest FAA score compared to existing fair FL methods. We note that fair FL methods (i.e., q-FFL~\citep{Li2019fair} and AFL~\citep{Mohri2019AFL}) have lower FAA scores than FedAvg, but their average test accuracy is worse. This is mainly because they mainly aim to improve bad agents (i.e., with high training loss), thus sacrificing the accuracy of agents with high-quality data. 

\begin{table}[h]
\caption{Comparison of FOCUS and the existing fair federated learning algorithms on the rotated MNIST dataset.}
\label{table:MNIST-ffl}
    \centering
    \begin{tabular}{lcccccccc}
    \toprule
        & \multirow{2}{*}{FOCUS} & \multirow{2}{*}{FedAvg} & \multicolumn{5}{c}{q-FFL} & AFL\\\cline{4-8}\\[-10pt]
        & & & $q=0.1$ & $q=1$ & $q=3$ & $q=5$ & $q=10$ & $\lambda=0.01$ \\
        \midrule
        Avg test accuracy & \textbf{0.953} & 0.929 & 0.922& 0.861 & 0.770 & 0.731 & 0.685 & 0.885\\
        Avg test loss & \textbf{0.152} & 0.246 & 0.269 & 0.489 & 0.777 & 0.900 & 1.084 & 0.429\\
        FAA & \textbf{0.094} & 0.363 & 0.388 & 0.612 & 0.547 & 0.419 & 0.253 & 0.220 \\
    \bottomrule
    \end{tabular}
\end{table}

\begin{table}[h]
\caption{Comparison of FOCUS and the existing fair federated learning algorithms on the rotated CIFAR dataset.}
    \label{table:CIFAR-ffl}
    \centering
    \begin{tabular}{lcccccccc}
    \toprule
        & \multirow{2}{*}{FOCUS} & \multirow{2}{*}{FedAvg} & \multicolumn{5}{c}{q-FFL} & AFL\\\cline{4-8}\\[-10pt]
        & & & $q=0.1$ & $q=1$ & $q=3$ & $q=5$ & $q=10$ & $\lambda=0.01$ \\
        \midrule
        Avg test accuracy  &\bf 0.688 & 0.654 & 0.648 & 0.592 & 0.426  & 0.181  & 0.121 & 0.661 \\ 
        Avg test loss & \textbf{1.133} & 2.386& 1.138 & 1.141 & 1.605 & 2.4746 & 2.526 & 1.666  \\
        FAA & 0.360 & 1.115 & 0.620 & 0.473 & 0.384 & \textbf{0.313} & 0.379 & 0.595\\
    \bottomrule
    \end{tabular}
\end{table}

\subsection{Comparison with state-of-the-art FL methods}
\label{append:SOTA}
We compare FOCUS with other SOTA FL methods, including FedMA \citep{wang2020federated}, Bayesian nonparametric FL \citep{yurochkin2019bayesian} and FedProx \citep{li2020federated}.
Specifically, the matching algorithm in \citep{yurochkin2019bayesian} is designed for only fully-connected layers, and the matching algorithm in \citep{wang2020federated} is designed for fully-connected and convolutional layers, while our experiments on CIFAR use ResNet-18 where the batch norm layers and residual modules are not considered in \citep{wang2020federated,yurochkin2019bayesian}. Therefore, we evaluate \citep{li2020federated,wang2020federated,yurochkin2019bayesian} on MNIST with a fully-connected network, and \citep{li2020federated} on CIFAR with a ResNet-18 model. 

The results on MNIST and CIFAR in \cref{table:MNIST-SOTA,table:CIFAR-SOTA} show that FOCUS achieves the highest average test accuracy and lowest FAA score than SOTA FL methods.


\begin{table}[h]
\caption{Comparison of FOCUS and other SOTA federated learning algorithms on the rotated MNIST dataset.}
\label{table:MNIST-SOTA}
    \centering
    \begin{tabular}{lccccccc}
    \toprule
        & \multirow{2}{*}{FOCUS} & \multirow{2}{*}{FedAvg} & \multicolumn{3}{c}{FedProx} & \multirow{2}{*}{FedMA} & Bayesian\\\cline{4-6}\\[-10pt]
        & & & $\mu =1$ & $\mu=0.1$ & $\mu=0.01$ &  & Nonparametric \\
        \midrule
        Avg test accuracy  & \textbf{0.953} & 0.929 & 0.908 & 0.927 & 0.929 & 0.753 & 0.517\\
        Avg test loss & \textbf{0.152} & 0.246 & 0.315 & 0.252 & 0.246 & 0.856 & 2.293\\
        FAA & \textbf{0.094} & 0.363 & 0.526 & 0.378 & 0.365 & 1.810 & 0.123\\
    \bottomrule
    \end{tabular}
\end{table}

\begin{table}
\caption{Comparison of FOCUS and other SOTA federated learning algorithms on the rotated CIFAR dataset.}
\label{table:CIFAR-SOTA}
    \centering
    \begin{tabular}{lccccc}
    \toprule
        & \multirow{2}{*}{FOCUS} & \multirow{2}{*}{FedAvg} & \multicolumn{3}{c}{FedProx}\\\cline{4-6}\\[-10pt]
        & & & $\mu =1$ & $\mu=0.1$ & $\mu=0.01$ \\
        \midrule
        Avg test accuracy  & \textbf{0.688} & 0.654 & 0.647 & 0.643 & 0.653\\
        Avg test loss & \textbf{1.133} & 2.386 & 1.206 & 2.151 & 2.404\\
        FAA & \textbf{0.360} & 1.115 & 0.397 & 0.884 & 0.787\\
    \bottomrule
    \end{tabular}
\end{table}

\subsection{Scalability with more agents}
\label{append:scale}
To study the scalability of \name, we evaluate the performance and fairness of \name and existing methods under 100 clients on \mnist. 
\cref{table:compare_100cli_mnist} shows that \name achieves the best fairness measured by \fairname and Agnostic Loss,  higher test accuracy, and lower test loss  than Fedavg and existing fair FL methods.

\begin{table}[ht]
    \centering
    \vspace{-3pt}
    \caption{ Comparison of different methods on \mnist 100 clients setting, in terms of average test accuracy (Avg Acc), average test loss (Avg Loss),  fairness \fairname  and existing fairness metric Agnostic loss. \name achieves the best fairness measured by \fairname.
    }
    \vspace{-3pt}
    \label{table:compare_100cli_mnist}
    \resizebox{1\linewidth}{!}{
    \begin{tabular}{llcccccccc}
    \toprule
      && \multirow{2}{*}{FOCUS} & \multirow{2}{*}{FedAvg} & \multicolumn{1}{c}{q-FFL} & AFL  & {Ditto} &{CGSV}\\\cmidrule(lr){5-8}\\[-10pt]
     && & &  $q=1$  & $\lambda=0.01$ &  {$\lambda=1$} &{$\beta=1$}  \\
    \midrule
    \multirow{3}{*}[-2pt]{Rotated MNIST (100 clients)}  & Avg Acc                  & \textbf{0.9533}           & 0.9236                     & 0.8371                       & 0.8813                  & 0.9351                    & 0.8691                   \\
 & Avg Loss                & \textbf{0.157}            & 0.2571                     & 0.5668                       & 0.4355                  & 0.2206                    & 0.6294                   \\
 & FAA                           & \textbf{0.5605}           & 1.0652                     & 1.5055                       & 0.8901                  & 0.7459                    & 1.2935                   \\
 & Agnostic Loss  & \textbf{0.5028}           & 0.8894                     & 1.4227                       & 0.7767                  & 0.620 &1.5133     \\
     \bottomrule
    \end{tabular}
    }
    \vspace{-3pt}
\end{table}

\subsection{Convergence of \name}
\label{append:convergence}
\begin{figure}[H]
    \centering
    \begin{subfigure}[b]{\textwidth}
        \centering
        \includegraphics[width=0.475\linewidth]{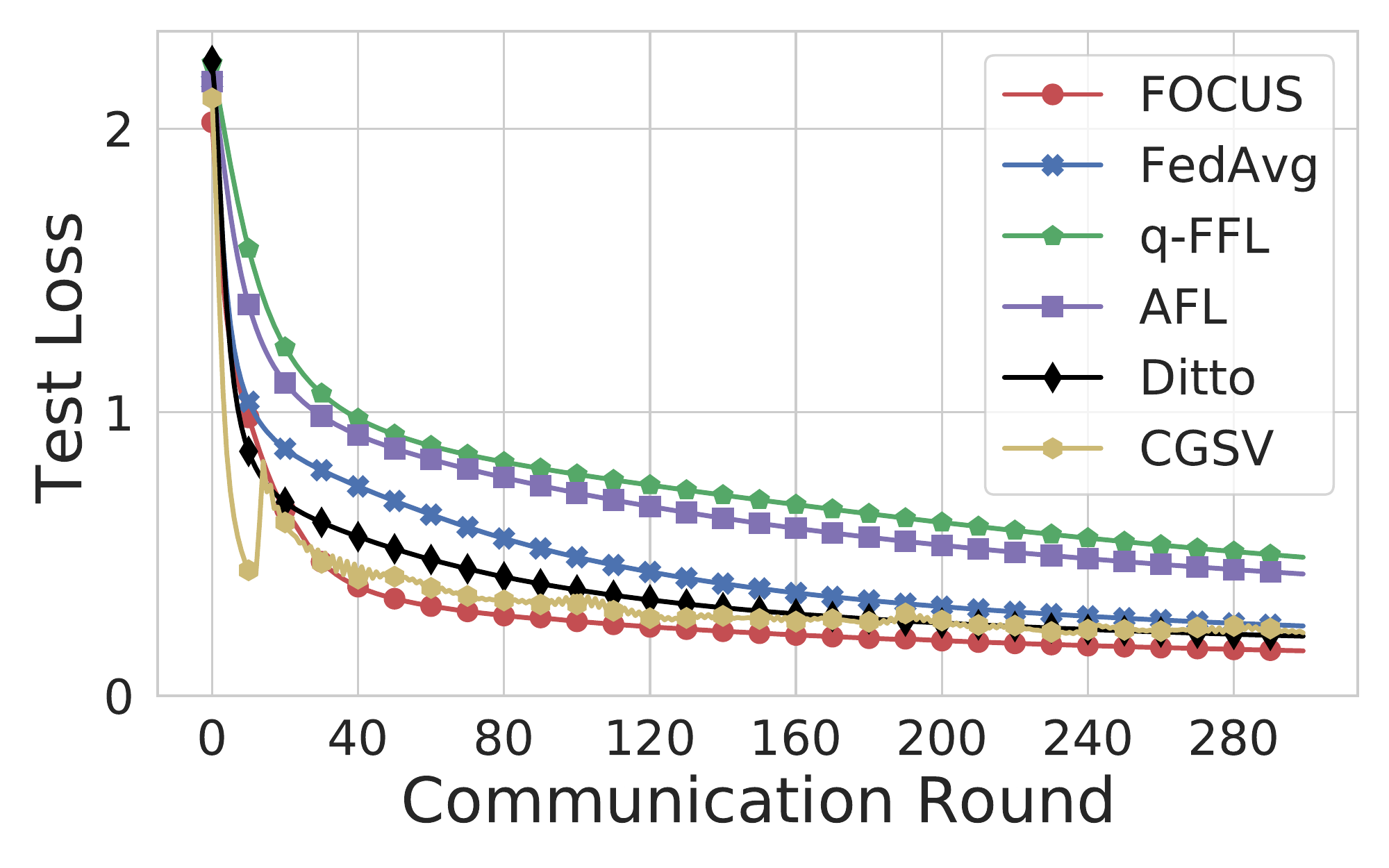}
        \hfill
        \includegraphics[width=0.475\linewidth]{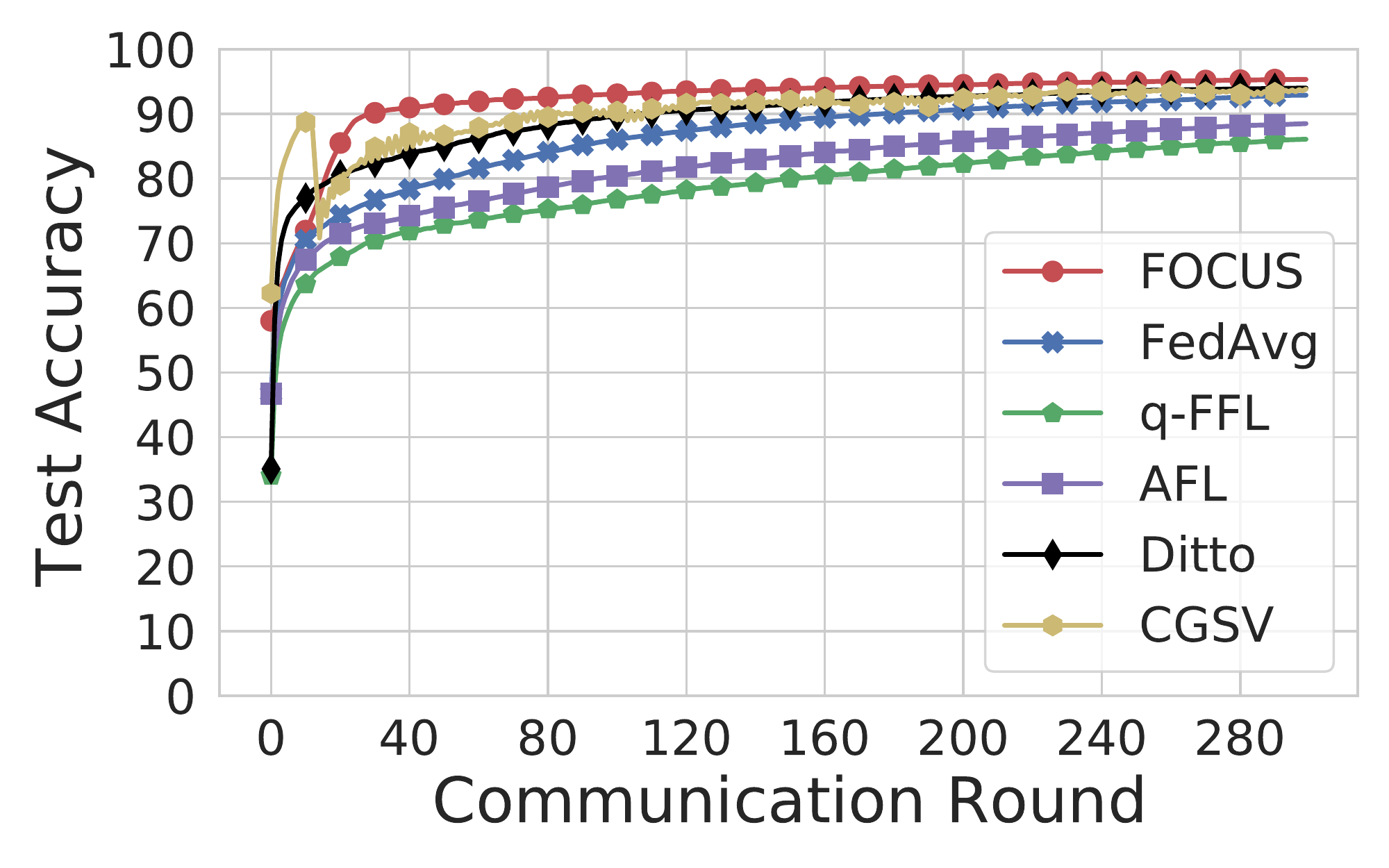}
        \caption{Rotated \mnist with 10 clients.}
    \end{subfigure}
    \vskip\baselineskip
    \begin{subfigure}[b]{\textwidth}
        \centering
        \includegraphics[width=0.475\linewidth]{figures/100cli_3cluster_test_loss.pdf}
        \hfill
        \includegraphics[width=0.475\linewidth]{figures/100cli_3cluster_test_acc.pdf}
        \caption{Rotated \mnist with 100 clients.}
    \end{subfigure}
    \vskip\baselineskip
    \begin{subfigure}[b]{\textwidth}
        \centering
      \includegraphics[width=0.475\linewidth]{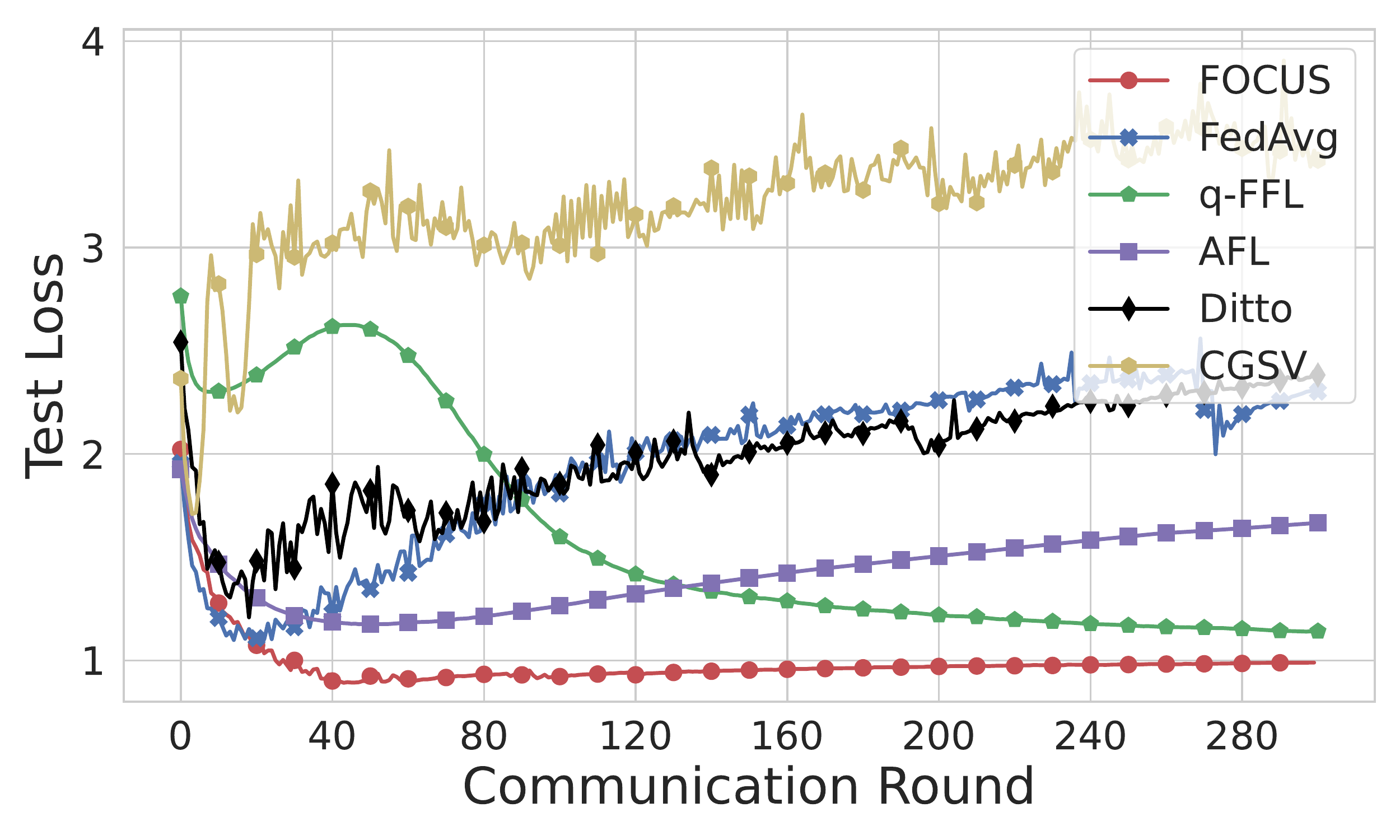}
        \hfill
        \includegraphics[width=0.475\linewidth]{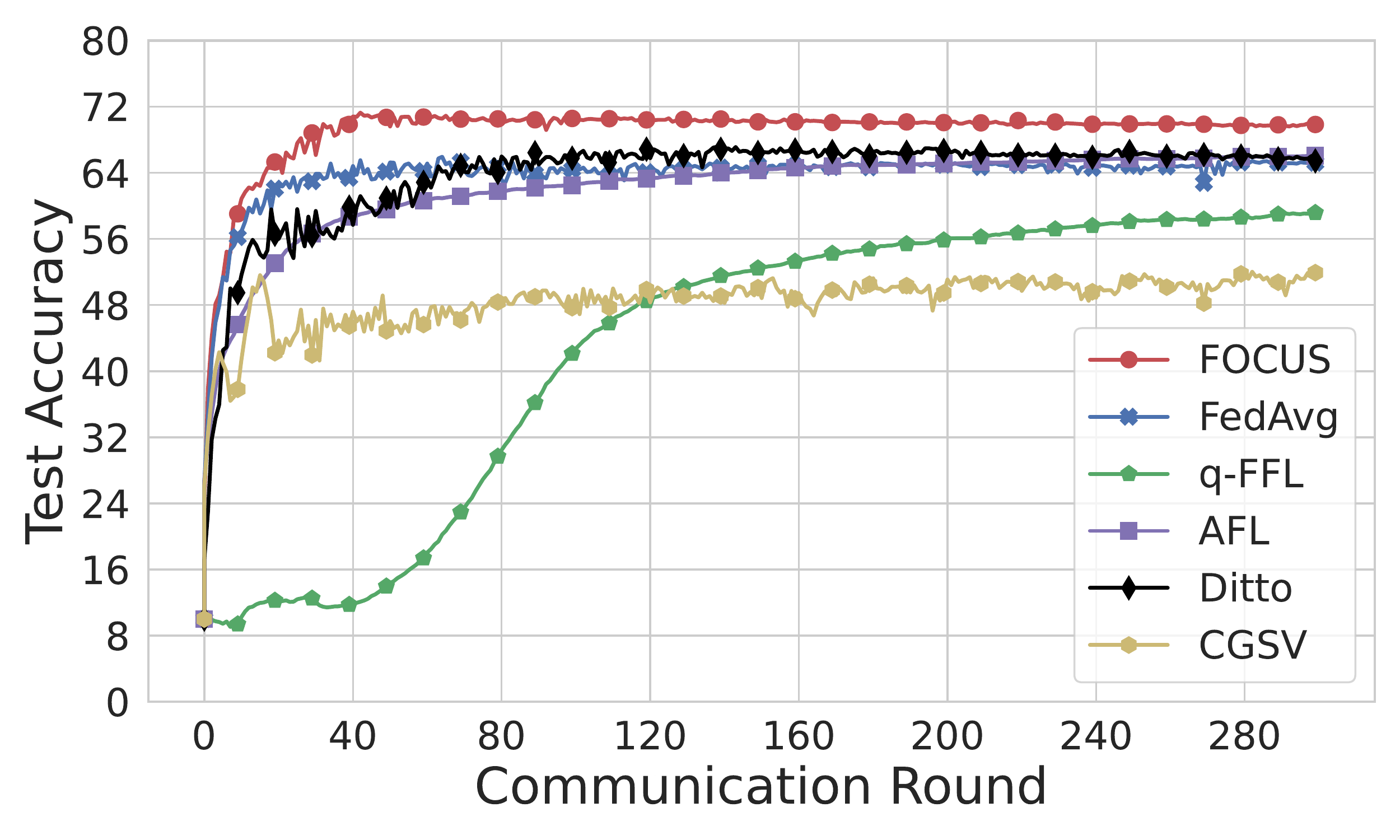}
        \caption{Rotated \cifar with 10 clients.}
    \end{subfigure}
    \caption{ The test accuracy and test loss of different methods over FL communication rounds on different datasets. \name converges faster and achieves higher accuracy and lower loss than other methods.}
    \label{fig:converge_compare}
\end{figure}

We report the test accuracy and test loss of different methods over FL communication rounds on Rotated \mnist with 10/100 clients and Rotated \cifar in \cref{fig:converge_compare}. The results show that \name converges faster and achieves higher accuracy and lower loss than other methods on both settings.

\subsection{Runtime analysis}
\label{append:runtime}
\paragraph{Computation time analysis for proposed  metric FAA and its scalability to more clients.}
In FAA, to calculate the maximal difference of excess risks for any pair of agents, it suffices to calculate the difference between the maximal per-client excess risks and the minimum per-client excess risk, and we don’t need to calculate the difference for any pairs of agents.
We compare the computation time (averaged over 100 trials) of FAA and existing fairness criteria (i.e., Accuracy Parity~\citep{Li2019fair}  and Agnostic Loss ~\citep{Mohri2019AFL}  ) under 10 clients and 100 clients on MNIST. \cref{tb:computation_metric}  shows that the computation of FAA is efficient even with a large number of agents.  Moreover, calculating the difference between maximal excess risk and minimum excess risk (i.e., FAA) is even faster than calculating the standard deviation of the accuracy between agents  (i.e., Accuracy Parity).

\begin{table}[]
\centering
  \caption{Computation time of different FL fairness metric on Rotated \mnist. The computation of \fairname is efficient under 10 clients and 100 client settings.
    }
    \label{tb:computation_metric}
\begin{tabular}{lll}
\toprule
& 10 clients                & 100 clients     \\ \midrule
Accuracy Parity~\citep{Li2019fair}  & 4.70 e-05 second          & 6.48 e-05 second \\ \midrule
Agnostic loss~\citep{Mohri2019AFL}          & {9.41 e-07 second} & 3.92 e-06 second \\ \midrule
FAA      & {6.09 e-06 second} & 4.27 e-05 second \\
\bottomrule
\end{tabular}
\end{table}

\paragraph{Communication rounds analysis.}
Here, we report the number of communication rounds that each method takes to achieve targeted accuracy on MNIST and CIFAR in \cref{tb:rounds_for_acc_mnist}. We note that \name requires significantly a smaller number of communication rounds than FedAvg, q-FFL, and AFL on both datasets, which demonstrates the small costs required by \name.

\begin{table}[]
\centering
  \caption{The number of communication rounds that different methods take to reach a target accuracy on Rotated \mnist. \name requires  a significantly smaller number of communication rounds than other methods.}
    \label{tb:rounds_for_acc_mnist}
\begin{tabular}{lllll}
\toprule
   & 70\%        & 80\%         & 85\%         & 90\%         \\\midrule
\name  & \textbf{9} & \textbf{16} & \textbf{20} & \textbf{29} \\\midrule
FedAvg & 10         & 51          & 88          & 177         \\\midrule
q-FFL  & 28         & 151         & 261         & $>$ 300        \\\midrule
AFL    & 16         & 94          & 180         & $>$ 300       \\
 \bottomrule
\end{tabular}
\end{table}

\begin{table}[]
\centering
  \caption{The number of communication rounds that different methods take to reach a target accuracy on Rotated \cifar. \name requires  a significantly smaller number of communication rounds than other methods.}
    \label{tb:rounds_for_acc_cifar}
\begin{tabular}{lllll}
\toprule
   & 55\%        & 60\%         & 65\%         & 70\%         \\\midrule
\name  & \textbf{8} & \textbf{10} & \textbf{19} & \textbf{37} \\\midrule
FedAvg & \bf 8         & 14          & 34         & $>$ 300         \\\midrule
q-FFL  & 182         &  $>$ 300       & $>$ 300         & $>$ 300        \\\midrule
AFL    & 24         & 53         & 190         & $>$ 300       \\
 \bottomrule
\end{tabular}
\end{table}

\paragraph{Training time and inference time analysis.}

In terms of runtime, we report the training time for one FL round (averaged over 20 trials) as well as inference time (averaged over 100 trials) in \cref{tb:train_infer_time}. 
Since the local updates and sever aggregation for different cluster models can be run in parallel, we find that FOCUS has a similar training time compared to FedAvg, q-FFL, and AFL which train one global FL model. 
For the inference time, FOCUS is slightly slower than existing methods by about 0.17 seconds due to the ensemble prediction of all cluster models at each client. However, we note that such cost is small and the forward passes of different cluster models for the ensemble prediction can also be made in parallel to further reduce the inference time.

\begin{table}[]
\centering
\caption{Training time per FL round and inference time for different methods on Rotate \mnist.}
\label{tb:train_infer_time}
\begin{tabular}{cccc}
\toprule
       & Training time per FL round  & Inference time       \\\midrule
\name   & {6.59}   second         & {0.28} second\\\midrule
FedAvg & 6.23   second                & 0.12   second\\\midrule
q-FFL  & 6.32      second             & 0.11   second     \\\midrule
AFL    & 6.24     second               & 0.12    second \\
 \bottomrule
\end{tabular}
\end{table}

\subsection{Comparison to FedAvg with clustering}
\label{append:hard_clustering}
In this section, we construct a new method by combining the clustering and Fedavg together (i.e., FedAvg-HardCluster), which serves as a strong baseline. 
Specifically, FedAvg-HardCluster works as below: 
\begin{itemize}[noitemsep,topsep=1pt,leftmargin=*]
    \item Step 1: before training, for each agent, it takes the arg max of the learned soft cluster assignment from FOCUS to get the hard cluster assignment (i.e., each agent only belongs to one cluster). 
    \item Step 2: during training, each cluster then trains a FedAvg model based on corresponding agents. 
    \item Step 3: during inference, each agent only uses the corresponding one cluster FedAvg model for inference. 
\end{itemize}

To compare the performance between FOCUS and FedAvg-HardCluster, we consider two scenarios on MNIST:
\begin{itemize}[noitemsep,topsep=1pt,leftmargin=*]
\item \textbf{Scenario 1}: underly clusters are clearly separatable, where each cluster contains samples from one distribution, which is the setting used in our paper. 
\item  \textbf{Scenario 2}: underlying clusters are not separatable, where each cluster has 80\%, 10\%, and 10\% samples from three different distributions, respectively. For example, the first underlying cluster contains 80\% samples without rotation, 10\% samples rotating 90 degrees, and 10\% samples rotating 180 degrees. 
\end{itemize} 
We observe that the learned soft cluster assignments from FOCUS align with the underlying distribution, so the hard cluster assignment for Step 1 in FedAvg-HardCluster is equal to the underlying ground-truth clustering for both scenarios.  

\cref{tb:compare_fedavg_cluster} presents the results of \name and FedAvg-HardCluster on Rotated \mnist under two scenarios.
\textbf{Under Scenario 1}, the accuracy of FOCUS and FedAvg-HardCluster is similar, and FOCUS achieves better fairness in terms of FAA. The results show that the hard clustering for FedAvg-HardCluster is as good as the soft clustering for FOCUS when the underlying clusters are clearly separable, which verifies that clustering is one of the key steps in FOCUS, and it aligns with our hypothesis for fairness under heterogeneous data. 
\textbf{Under Scenario 2}, FOCUS achieves higher accuracy and better FAA fairness than FedAvg-HardCluster. The results show that when underly clusters are not separatable, soft clustering is better than hard clustering since each agent can benefit from multiple cluster models with the soft $\pi$ learned from the EM algorithm in FOCUS. 

\begin{table}[]
\centering
\caption{Comparison between \name and FedAvg-HardCluster on Rotate \mnist under two scenarios.}
\label{tb:compare_fedavg_cluster}
\begin{tabular}{lcc|cc}
\toprule
& \multicolumn{2}{c}{\begin{tabular}[c]{@{}c@{}}Scenario 1 \\ (underly clusters are clearly separatable)\end{tabular}} & \multicolumn{2}{c}{\begin{tabular}[c]{@{}c@{}}Scenario 2\\  (underly clusters are not separatable)\end{tabular}} \\\midrule
& FOCUS                                                   & FedAvg-HardCluster                                         & FOCUS                                                 & FedAvg-HardCluster                                       \\\midrule
Avg test acc  & 0.953                                                   & 0.954                                                      & \textbf{0.814}                                        & 0.812                                                    \\\midrule
Avg test loss & 0.152                                                   & 0.152                                                      & \textbf{1.168}                                        & 1.244                                                    \\\midrule
FAA           & \textbf{0.094}                                          & 0.099                                                      & \textbf{0.449}                                        & 0.459                                                    \\\midrule
Agnostic loss & 0.224                                                   & 0.224                                                      & \textbf{1.333}                                        & 1.397   \\  \bottomrule                                  
\end{tabular}
\end{table}

\subsection{Effect of the number of the clusters $M$}
The performance of FOCUS would not be harmed if the selected number of clusters is larger than the number of underlying clusters since the superfluous clusters would be useless (the corresponding soft cluster assignment $\pi$ goes to zero). On the other hand, when the selected number of clusters is smaller than the number of underlying clusters, FOCUS would converge to a solution when some clusters contain agents from more than one underlying cluster.  

Empirically, in \cref{tb:different_m_mnist}, we have 3 true underlying clusters while we set $M=1,2,3,4$ in our experiments, and we see that when $M=3$ and $M=4$, FOCUS achieves similar accuracy and fairness, which verifies our hypothesis that the superfluous clusters would become useless. When $M=2$, FOCUS even achieves the highest fairness, which might be because one cluster benefits from the shared knowledge of multiple underlying clusters. When $M=1$, FOCUS reduces to FedAvg, which does not have the clustering mechanism, leading to the lowest accuracy and fairness under heterogeneous data. 

\begin{table}[]
\centering
\caption{The effect of $M$ on Rotate \mnist when the number of underlying clusters is 3. }
\label{tb:different_m_mnist}
\begin{tabular}{lccccc}\toprule
  & M=1    & M=2             & M=3             & M=4             \\\midrule
Avg test acc       &0.929           & 0.952          &  \bf 
 0.953          &  \bf 0.953 \\\midrule
Avg test loss           & 0.246      & 0.167          & \textbf{0.152} & 0.153          \\\midrule
FAA           &0.363                & \textbf{0.079} & 0.094          & 0.091          \\\midrule
Agnostic loss & 0.616 & 0.272          & 0.224          & \textbf{0.223} \\ \bottomrule   
\end{tabular}
\end{table}

\begin{table}[]
    \centering
    \caption{The effect of $M$ on Rotate \cifar when the number of underlying clusters is 2.}
    \begin{tabular}{lccccc}
    \toprule
         & M=1 & M=2 & M=3 & M=4 \\
         \midrule
         Avg test acc & 0.654 & 0.688 & \textbf{0.696} & 0.693\\
         Avg Loss & 2.386 & 1.133 & 0.932 & \textbf{0.921}\\
         FAA & 1.115 & 0.360 & \textbf{0.323} & 0.350 \\
         Agnostic loss & 3.275 & 1.294 & 1.115 & \textbf{1.098}\\
        \bottomrule
    \end{tabular}
    \label{tab:different_m_cifar}
\end{table}

\subsection{Histogram of loss on CIFAR}
\label{append:CIFAR-loss}
\begin{figure}[H]
     \centering
     \begin{subfigure}[b]{0.5\textwidth}
         \centering
         \includegraphics[width=\textwidth]{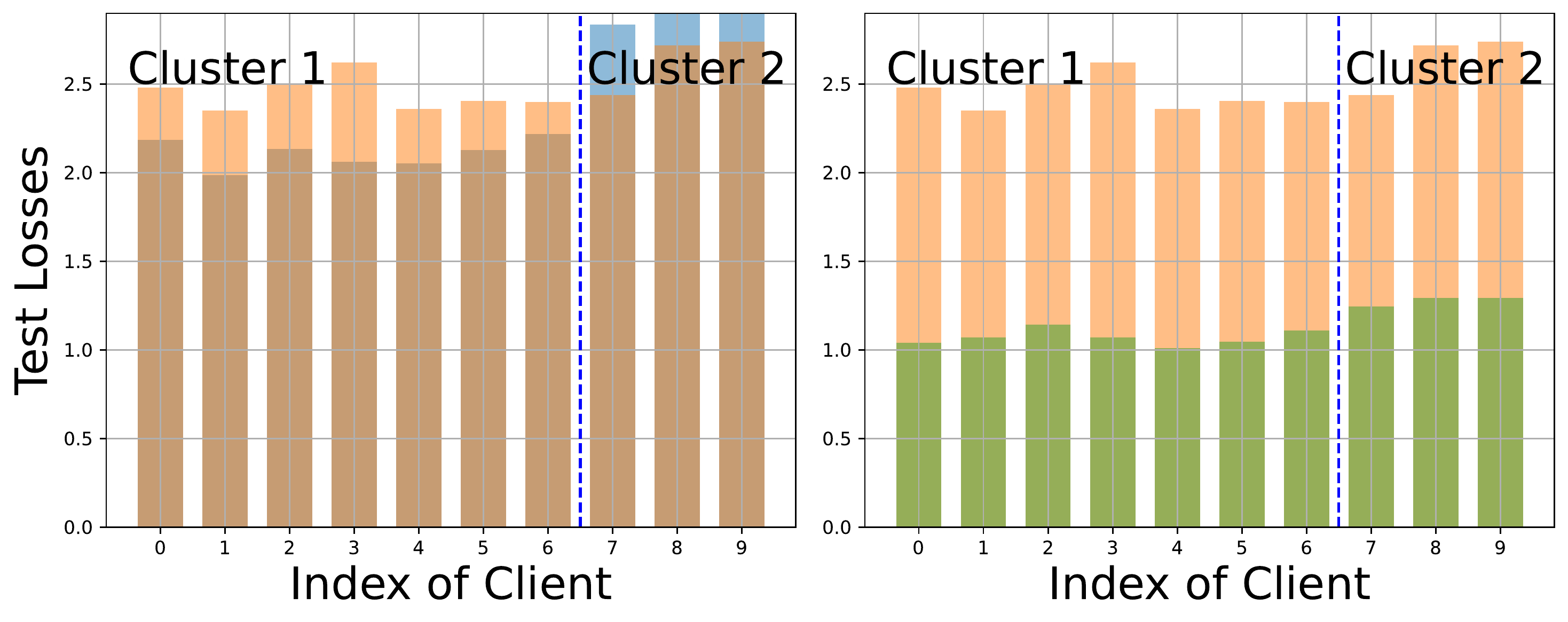}
     \end{subfigure}
        \caption{\small The excess risk of different agents trained with FedAvg (left) and \name (right) on \cifar dataset.}
        \label{fig:excess_risk_cifar}
\end{figure}

\cref{fig:excess_risk_cifar} shows the surrogate excess risk of every agent trained with FedAvg and \name on \cifar dataset. For the outlier cluster that rotates 180 degrees (i.e., 2rd cluster), FedAvg has the highest test loss for the 9th agent, resulting in a high excess risk of 2.74. In addition, the agents in 1st cluster trained by FedAvg are influenced by the FedAvg global model and have high excess risk. On the other hand,  \name successfully identifies the outlier distribution in 2nd cluster, leading to a much lower excess risk among agents with a more uniform excess risk distribution. 
Notably, \name reduces the surrogate excess risk for the 9th agent to 0.44.

\section{Convergence Proof}
\label{appendixA}
\subsection{Convergence of Linear Models (\cref{theorem:linear_converge})}
\label{appendixA1}

\subsubsection{Key Lemmas}
We need to state two lemmas first before proving \cref{theorem:linear_converge}.
\begin{lemma}
\label{lemma:1}
Suppose $e\in S_m$ and the $m$-th cluster is the one closest to $w^*_m$. Assume $\|w_m^{(\revise{t})} - w_m^*\| \leq \alpha < \beta \leq \min_{m^\prime \neq m}\|w_{m^\prime}^{\revise{(t)}} - w^*_m\|$. Then the E-step updates as
\begin{equation}
    \pi_{em}^{\revise{(t+1)}} \geq \frac{\pi_{em}^{\revise{(t)}}}{\pi_{em}^{\revise{(t)}} + (1 - \pi_{em}^{\revise{(t)}})\exp\Big(-(\beta^2 - \alpha^2 - 2(\alpha + \beta)r )\delta^2\Big)}
\end{equation}
\end{lemma}

\noindent\textbf{Remark.} Our assumption of proper initialization guarantees that $\|w_m^{(0)} - w^*_m\| \leq \alpha$ while $\forall m^\prime$, we have $\|w_{m^\prime} - w^*_m\|_2 \geq \|w^*_m - \mu_{m^\prime}^*\| - \|w_{m^\prime} - \mu_{m^\prime}^*\| = R - \alpha$. Hence, we substitute $\beta = R - \alpha$ and $\alpha = \frac{R}{2} - r - \Delta$, which yields
\begin{equation}
    \pi_{em}^{\revise{(t+1)}} \geq \frac{\pi_{em}^{\revise{(t)}}}{\pi_{em}^{\revise{(t)}} + (1-\pi_{em}^{\revise{(t)}}) \exp(-2R\Delta \delta^2)}, \quad \forall e\in S_m
\end{equation}

For M-steps, the local agents are initialized with $\theta^{(0)}_{em} = w_m^{\revise{(t)}}$. Then for $\revise{k}=1,\dots, \revise{K}-1$, each agent use local SGD to update its personal model:
\begin{equation}
    \theta_{em}^{\revise{(k+1)}} = \theta_{em} - \eta_{\revise{k}} g_{em}(\theta_{em}) = \theta_{em}^{\revise{(k)}} - \eta_{\revise{k}} \nabla\sum_{i=1}^{n_e} \ell(h_{\theta_{em}}(x_e^{(i)}),y_e^{(i)}).
\end{equation}

To analyze the aggregated model \cref{M:agg}, we define a sequence of virtual aggregated models $\hat w_m^{\revise{(k)}}$.
\begin{equation}
\label{eq:virtual}
    \hat w_m^{\revise{(k)}} = \sum_{e=1}^E\frac{\pi_{em}\theta_{em}^{\revise{(k)}}}{\sum_{e^\prime = 1}^E \pi_{e^\prime m}}.
\end{equation}

\begin{lemma}
\label{lemma:2}
Suppose any agent $e\in S_m$ has a soft clustering label $\pi_{em}^{\revise{(t+1)}} \geq p$. Then one step of local SGD updates $\hat w_m^{\revise{(k)}}$ by \cref{eq:lemma:2}, if the learning rate $\eta_{\revise{k}} \leq \frac{1}{4\delta^2}$.

\begin{equation}
\label{eq:lemma:2}
    \mathbb E \|\hat w_m^{\revise{(k+1)}} - w^*_m\|_2^2 \leq (1 - 2\eta_{\revise{k}}\gamma_m p \delta^2) \mathbb E\|\hat w_m^{\revise{(k+1)}} - w^*_m\|_2^2 + \eta_{\revise{k}} A_1 + \eta_{\revise{k}}^2 A_2.
\end{equation}
\begin{equation}
    A_1 = 4\gamma_m r\delta^2  + 2\delta^2E(1-p), A_2 = 16E(\revise{K} - 1)^2\delta^4 + O(\frac{d}{n_e}) E (\delta^4 +\delta^2 \sigma^2)
    \end{equation}
\end{lemma}

\textbf{Remark.} Using the recursive relation in \cref{lemma:2}, if the learning rate $\eta_{\revise{k}}$ is fixed, the sequence $\hat w_m^{\revise{(k)}}$ has a convergence rate of
\begin{equation}
    \mathbb E\|\hat w_m^{\revise{(k)}} - w^*_m\|_2^2 \leq (1 - 2\eta \gamma_m p\delta^2)^{\revise{k}} \mathbb E\|\hat w_m^{(0)} - w^*_m\|_2^2 + \eta \revise{k} (A_1 + \eta A_2).
\end{equation}

\subsubsection{Completing the Proof of \cref{theorem:linear_converge}}
We now combine \cref{lemma:1} and \cref{lemma:2} to prove \cref{theorem:linear_converge}. The theorem is restated below.

\begingroup
\def\thetheorem{\ref{theorem:linear_converge}}
\begin{theorem}
With the assumptions 1 and 2, $n_e = O(d)$, if learning rate $\eta\leq \min(\frac{1}{4\delta^2}, \frac{\beta}{\sqrt{\revise{T}}})$, 
\small
\begin{align}
    &\pi_{em}^{\revise{(T)}} \geq \frac{1}{1+(M-1)\cdot\exp(-2R\delta^2 \Delta_0 K)}, \forall e\in S_m\label{eq:pi_}\\
    &\mathbb E\|w_m^{\revise{(T)}} - w^*_m\|_2^2 \leq (1 - \frac{2\eta \gamma_m\delta^2}{M})^{KT} (\|w_m^{(0)} - w^*_m\|_2^2 + A)
    +2M\revise{K}r
    + \frac{M\delta^2 E\beta}{2\sqrt{\revise{T}}} O(\revise{K}^3, \sigma^2).\label{eq:w_}
\end{align}
\normalsize
where $K$ is the total number of communication rounds; $T$ is the number of iterations each round; $\gamma_m = |S_m|$ is the number of agents in the $m$-th cluster, and

\vspace{-12pt}
\small
\begin{equation}
   A= \frac{2E\revise{K}(M-1)
   \delta^2}{(1 - \frac{2\eta \delta^2 \gamma_m}{M})^{\revise{K}} - \exp(-2R\delta^2 \Delta_0)}.
\end{equation}
\normalsize
\end{theorem}
\addtocounter{theorem}{-1}
\endgroup

\begin{proof}
We prove \cref{theorem:linear_converge} by induction. Suppose
\begin{align}
    &\pi_{em}^{\revise{(t)}} \geq \frac{1}{1 + (M-1) \exp(-2R\delta^2 \Delta_0 \revise{t})}\\
    &\mathbb E\|w_m^{\revise{(t)}} - w^*_m\|^2 \leq (1- \frac{2\eta\gamma_m \delta^2}{M})^{\revise{Kt}}(\|w_m^{(0)} - w^*_m\|^2) + A\Big((1- \frac{2\eta\gamma_m \delta^2}{M})^{\revise{Kt}} - \exp(-2R\delta^2 \Delta_0 \revise{t})\Big)  \nonumber\\
    &\hspace{25em}+\frac{\eta B} {1 - (1 - \frac{2\eta \gamma_m\delta^2}{M})^{\revise{K}}} .\label{eq:linear_M}
\end{align}
where $B = [16 E \delta^4 \revise{K}^3+ E\revise{K}(\delta^4 + \delta^2\sigma^2)]\eta + 4\gamma_m r \delta^2 \revise{K}.$

Then according to \cref{lemma:1},
\begin{align}
    \pi_{em}^{\revise{(t+1)}} &\geq \frac{\pi_{em}^{\revise{(t)}}}{\pi_{em}^{\revise{(t)}} + (1-\pi_{em}^{\revise{(t)}}) \exp(-2R\Delta_0 \delta^2)}\\
    &\geq \frac{1}{1+(M-1)\exp(-2R \delta^2 \Delta_0 \revise{t}) \exp(-2R\Delta_0\delta^2)}\\
    &\geq \frac{1}{1+(M-1)\exp(-2R\Delta_0\delta^2(\revise{t}+1))}.
\end{align}

We recall the virtual sequence of $\hat w_m$ defined by \cref{eq:virtual}. Since models are synchronized after $\revise{K}$ rounds, the know $\hat w_m^{(0)} = w_m^{\revise{(t)}}$ and $w_m^{\revise{(t+1)}} = \hat w_m^{\revise{(K)}}$. We then apply \cref{lemma:2} to prove the induction. Note that instead of proving \cref{eq:w_}, we prove a stronger induction hypothesis of \cref{eq:linear_M}.

\begin{align}
    &\mathbb E\|w_m^{\revise{(t+1)}} - w^*_m\|^2 \nonumber \\
    &= \mathbb E\|\hat w_m^{\revise{(K)}} - w^*_m\|^2 \\
    &\leq (1 - 2\eta \gamma_m p \delta^2)^{\revise{K}} \mathbb E\|\hat w_m^{\revise{(t)}} - w^*_m\|^2 + \eta \revise{K}(A_1 + \eta A_2)\\
    &\leq (1 - 2\eta \gamma_m p \delta^2)^{\revise{K}} \Big((1 - \frac{2\eta \gamma_m\delta^2}{M})^{\revise{Kt}}\|w_m^{(0)} - w^*_m\|^2 +  A ((1 - \frac{2\eta \gamma_m\delta^2}{M})^{\revise{Kt}} - \exp(-2R\Delta_0 \delta^2 \revise{t})) \nonumber\\
    & \hspace{5em}+\frac{\eta B}{1 - (1 - \frac{2\eta \gamma_m\delta^2}{M})^{\revise{K}}} \Big) +\eta \revise{K} (4\gamma_m r \delta^2 + 2\delta^2 E(1-p)) + \eta^2 \revise{K} A_2\\
    &\leq (1 - \frac{2\eta \gamma_m \delta^2}{M})^{(\revise{t}+1)\revise{K}} \|w_m^{(0)} - w^*_m\|^2 \nonumber\\
     &+\underbrace{A(1 - \frac{2\eta \gamma_m \delta^2}{M})^{(\revise{t}+1)\revise{K}} - A\exp(-2R\Delta_0 \delta^2 \revise{t})(1 - \frac{2\eta\gamma_m \delta^2}{M})^{\revise{K}} + 2\delta^2 E(1-p)}_{D_1}\nonumber\\
     &+\underbrace{(1 - \frac{2\eta \gamma_m \delta^2}{M})^{\revise{K}}  \frac{\eta B}{1 - (1 - \frac{2\eta \gamma_m\delta^2}{M})^{\revise{K}}} + 4\eta \revise{K}\gamma_m r \delta^2 + \eta^2 \revise{K}A_2}_{D_2}. \label{expand}
\end{align}

Note that $1-p \leq (M-1)\exp(-2R\Delta_0\delta^2 \revise{t})$, so
\begin{align}
    D_1 &\leq A(1 - \frac{2\eta \gamma_m \delta^2}{M})^{(\revise{t}+1)\revise{K}} - A \exp(-2R\Delta_0 \delta^2 \revise{t})(1 - \frac{2\eta \gamma_m \delta^2}{M})^{\revise{K}} + 2\delta^2 E\revise{K}(M-1)\exp(-2R\Delta_0 \delta^2 \revise{t})\nonumber\\
    &\leq A ((1 - \frac{2\eta \gamma_m\delta^2}{M})^{(\revise{t}+1)\revise{K}} - \exp(-2R\Delta_0 \delta^2 (\revise{t}+1)))  \label{D1}
\end{align}

For $D_2$ we have
\begin{align}
    D_2 &\leq (1 - \frac{2\eta \gamma_m \delta^2}{M})^{\revise{K}} \frac{\eta B} {[1 - (1 - \frac{2\eta \gamma_m\delta^2}{M})^{\revise{K}}]} + 4\eta \gamma_m r \delta^2 \revise{K} + 16 \eta^2 E \delta^4 \revise{K}^3 +  \eta^2 E\revise{K} O(\delta^4 + \delta^2 \sigma^2)\nonumber \\
    &= \frac{\eta B}{1 - (1 - \frac{2\eta\gamma_m\delta^2}{M})^{\revise{K}}}. \label{D2}
\end{align}

Finally we combine \cref{expand,D1,D2} so
{\small
\begin{align}
    \mathbb E\|w_m^{\revise{(t+1)}} - w^*_m\|^2 &\leq (1 - \frac{2\eta \gamma_m \delta^2}{M})^{(\revise{t}+1)\revise{K}}\|w_m^{(0)}- w^*_m\|^2 + A\Big((1 - \frac{2\eta \gamma_m \delta^2}{M})^{(\revise{t}+1)\revise{K}} - \exp(-2R\delta^2 \Delta_0 (\revise{t}+1))\Big)\nonumber\\
    & \hspace{20em} + \frac{\eta B}{1 - (1 - \frac{2\eta \gamma_m\delta^2}{M})^{\revise{K}}}. \label{eq:50_}
\end{align}
}

Since it is trivial to check that both induction hypotheses hold when $\revise{t} = 0$, the induction hypothesis holds. Note that
$\revise{K} \geq 1$, so
\begin{equation}
    \frac{\eta B}{1 - (1 - \frac{2\eta \gamma_m \delta^2}{M})^{\revise{K}}} \leq \eta B \frac{M}{2\eta \gamma_m \delta^2} \leq 2M\revise{K}r + \frac{M\delta^2 E \beta}{2\sqrt{\revise{T}}} O(\revise{K}^3, \delta^2). \label{eq:51}
\end{equation}
Combining \cref{eq:50_} and \cref{eq:51} completes our proof.
\end{proof}

\subsubsection{Deferred Proofs of Key Lemmas}
\textbf{Lemma 1.}
\begin{proof}
For simplicity, we abbreviate the model weights $w_m^{\revise{(t)}}$ by $w_m$ in the proof of this lemma.
The $n$-th E step updates the weights $\Pi$ by
\begin{equation}
    \pi_{em}^{\revise{(t+1)}} = \frac{\pi_{em}^{\revise{(t)}} \exp[-\mathbb E_{(x,y)\sim D_e}({w_m}^Tx - y)^2]} {\sum_{m^\prime} \pi_{em^\prime}^{\revise{(t)}} \exp[-\mathbb E_{(x,y)\sim D_e}({w_{m^\prime}}^T x - y)^2]}
\end{equation}
so
\begin{align}
    \pi_{em}^{\revise{(t+1)}} &=  \frac{\pi_{em}^{\revise{(t)}} \exp(-\|w_m^{\revise{(t)}} - \mu_e\|^2 \delta^2)}{\sum_{m^\prime} \pi^{\revise{(t)}}_{em^\prime} \exp[-\|{w_m^\prime}^{\revise{(t)}} - \mu_e\|^2\delta^2]}\\
    & \geq \frac{\pi_{em}^{\revise{(t)}} \exp(-(\beta - r)^2 \delta^2)}{\pi_{em}^{\revise{(t)}} \exp(-(\beta-r)^2\delta^2) +\sum_{m^\prime\neq m} \pi_{em^\prime}^{\revise{(t)}} \exp(-(\alpha+r)^2\delta^2)}\\
    &\geq \frac{\pi_{em}^{\revise{(t)}}}{\pi_{em}^{\revise{(t)}} + (1 - \pi_{em}^{\revise{(t)}})\exp\Big(-(\beta^2 - \alpha^2 - 2(\alpha + \beta)r )\delta^2\Big)}
\end{align}
\end{proof}

\textbf{Lemma 2.}
\begin{proof}
Notice that local datasets are generated by $X_e\sim \mathcal N(0,\delta^2 \mathbf 1^{n_e\times d})$ and $y_e = X_e \mu_e + \epsilon_e$ with $\epsilon_e \sim \mathcal N(0,\sigma^2)$. Therefore, 

\begin{align}
    \|\hat w_m^{\revise{(k+1)}} - w^*_m\|^2 &= \|w_m^{\revise{(k)}} - w^*_m - \eta_{\revise{k}} g_{\revise{k}}\|^2 \\
    &=\|\hat w_m^{\revise{(k)}} - w^*_m - \eta_{\revise{k}} \frac{2}{n_e} \sum_{e} \pi_{em} X_e^T X_e(\theta_{em}^{\revise{(k)}} - \mu_e) + \frac{2\eta_{\revise{k}}}{n_e}\sum_{e} \pi_{em} X_e^T \epsilon_e\|^2 \\
    &= \|\hat w_m^{\revise{(k)}} - w^*_m - \hat g_{\revise{k}}\|^2 +  \eta_{\revise{k}}^2 \|g_{\revise{k}} - \hat g_{\revise{k}}\|^2 + 2 \eta_{\revise{k}} \langle w_m^{\revise{(k)}} - w^*_m - \hat g_{\revise{k}}, \hat g_{\revise{k}} - g_{\revise{k}}\rangle. \label{eq:42}
\end{align}
where $\hat g_{\revise{k}} = \frac{2}{n_e}\sum_{e}\pi_{em} \mathbb E(X_e^T X_e)(\theta_{em}^{\revise{(k)}} - \mu)$. Since the expectation of the last term in \cref{eq:42} is zero, we only need to estimate the expectation of $\|\hat w_m^{\revise{(k)}} - w^*_m - \eta_{\revise{k}} \hat g_{\revise{k}}\|^2$ and $\|\hat g_{\revise{k}} - g_{\revise{k}}\|^2$.

\begin{align}
    &\|\hat w_m^{\revise{(k)}} - w^*_m - \eta_{\revise{k}} \hat g_{\revise{k}} \|^2 \nonumber \\
    &= \|\hat w_m^{\revise{(k)}} - w^*_m \|^2 +
    \frac{4\eta_{\revise{k}}^2}{n_e^2}\sum_{e} \pi_{em}\mathbb E(X_e^T X_e) \|\theta_{em}^{t} - \mu_e\|^2 - \frac{4\eta_{\revise{k}}}{n_e} \sum_{e}\pi_{em}\langle \hat w_m^{\revise{(k)}} - w^*_m, \mathbb E(X_e^T X_e)(\theta_{em}^{\revise{(k)}} - \mu_e)\rangle\nonumber \\
    &= \|\hat w_m^{\revise{(k)}} - w^*_m\|^2 + 4\eta_{\revise{k}}^2 \delta^2 \sum_{e}\pi_{em} \|\theta_{em}^{\revise{(k)}} - \mu_e\|^2 - \underbrace{4\eta_{\revise{k}} \langle \hat w_m^{\revise{(k)}} - w^*_m, \sum_{e}\pi_{em} \delta^2(\theta_{em}^{\revise{(k)}} - \mu_e)\rangle}_{C_1}.
\end{align}

\begin{align}
    C_1 &= -4\eta_{\revise{k}} \sum_{e}\pi_{em} \langle \hat w_m^{\revise{(k)}} - \theta_{em}^{\revise{(k)}}, \delta^2 (\theta_{em}^{\revise{(k)}} - \mu_e)\rangle - 4\eta_{\revise{k}} \sum_{e}\pi_{em}\langle \theta_{em}^{\revise{(k)}} - w^*_m, \delta^2 (\theta_{em}^{\revise{(k)}} - \mu_e)\rangle\\
    &\leq 4\sum_{e}\pi_{em} \|\hat w_m^{\revise{(k)}} - \theta_{em}^{\revise{(k)}}\|^2 + 4\delta^4 \eta_{\revise{k}}^2 \sum_{e}\pi_{em} \|\theta_{em}^{\revise{(k)}} - \mu_{e}\|^2 - 4\eta_{\revise{k}} \delta^2\sum_{e} \pi_{em} \|\theta_{em}^{\revise{(k)}} - \mu_e\|^2 \nonumber \\
    &\hspace{18em} - 4\eta_{\revise{k}} \delta^2 \underbrace{\sum_{e}\pi_{em} \langle \mu_e - w^*_m ,\theta_{em}^{\revise{(k)}} - \mu_e\rangle}_{C_2}
\end{align}

Since $\eta_{\revise{k}} \leq \frac{1}{4\delta^2}$,
\begin{align}
    &\mathbb E\|\hat w_m^{\revise{(k)}} - w^*_m - \eta_{\revise{k}} \hat g_{\revise{k}}\|^2 \\
    &\leq \mathbb E\|\hat w_m^{\revise{(k)}} - w^*_m\|^2+ (8\delta^4 \eta_{\revise{k}}^2 - 4\eta_{\revise{k}} \delta^2)\sum_{e} \pi_{em}\mathbb E\|\theta_{em}^{\revise{(k)}} - \mu_e\|^2 + 4\sum_{e}\pi_{em} \mathbb E\|\hat w_m^{\revise{(k)}} - \theta_{em}^{\revise{(k)}}\|^2+C_2\\
    &\leq \mathbb E\|\hat w_m^{\revise{(k)}} - w^*_m\|^2 -2\eta_{\revise{k}} \delta^2  \sum_{e}\pi_{em} \mathbb E\|\theta_{em}^{\revise{(k)}} - \mu_e\|^2 + 4\sum_{e}\pi_{em}\mathbb E\|\hat w_m^{\revise{(k)}} - \theta_{em}^{\revise{(k)}}\|^2 + C_2
\end{align}

Note that 
\begin{align}
    &\sum_{e}\pi_{em} \mathbb E \|\theta_{em}^{\revise{(k)}}  - \mu_e\|^2\\
    &= \sum_{e\in S_m} \pi_{em} \mathbb E\|\theta_{em}^{\revise{(k)}}  - \mu_e\|^2 + \sum_{e\not\in S_m} \pi_{em} \mathbb E\|\theta_{em}^{\revise{(k)}}  - \mu_e\|^2\\
    &\geq \sum_{e\in S_m} \pi_{em} (\mathbb E\|\theta_{em}^{\revise{(k)}}  - w^*_m\|^2 + 2r + r^2) + \sum_{e\not\in S_m}\pi_{em}\mathbb E\|\theta_{em}^{\revise{(k)}} - \mu_e\|^2 \\
    &= \sum_{e\in S_m} \pi_{em} (\mathbb E\|\hat w_m ^{\revise{(k)}} - w^*_m\|^2 +\mathbb E\|\hat w_m^{\revise{(k)}} - \theta_{em}^{\revise{(k)}} \|^2 + 2r + r^2) + \sum_{e\not\in S_m}\pi_{em}\mathbb E\|\theta_{em}^{\revise{(k)}} - \mu_e\|^2
\end{align}

And since $\hat w_m^{\revise{(k)}} = \mathbb E \sum_{e}\pi_{em} \theta_{em}^{\revise{(k)}}$, we have 
\begin{align}
    4\mathbb E\sum_{e}\pi_{em} \|\hat w_m^{\revise{(k)}} - \theta_{em}^{\revise{(k)}}\|^2 &\leq 4 \mathbb E\sum_{e}\pi_{em} \|\hat w_m^{(0)} - \theta_{em}^{\revise{(k)}}\|^2 \\
    &\leq 4\sum_{e} \pi_{em} (\revise{K}-1) \mathbb E \sum_{t^\prime}^{t-1} {\eta_{\revise{k}}^\prime}^2 \|\frac{2}{n_e} X_e^T X_e(\theta_{em}^{\revise{(k)}} - \mu_e)\|^2\\
    &\leq 16 \eta_{\revise{k}}^2 E(\revise{K}-1)^2 \delta^4.  
\end{align}

Thus,
\begin{align}
    \mathbb E\|\hat w_m^{\revise{(k)}} - w^*_m - \eta_{\revise{k}} \hat g_{\revise{k}}\|^2 &\leq (1 - 2\eta_{\revise{k}} \delta^2 \sum_{e} \pi_{em}) \mathbb E\|\hat w_m^{\revise{(k)}} - w^*_m\|^2  + 16 \eta_{\revise{k}}^2 E(\revise{K}-1)^2 \delta^4 \nonumber \\
    & \underbrace{- 2\eta_{\revise{k}} \delta^2 \sum_{e\not\in S_m}\pi_{em} \mathbb E\|\theta_{em}^{\revise{(k)}} - \mu_e\|^2 - 4\eta_{\revise{k}} \delta^2 \sum_{e}\pi_{em}\langle \theta_{em}^{\revise{(k)}} - \mu_e ,\mu_e - w^*_m\rangle}_{C_3}
\end{align}

Since
\begin{align}
    C_3 &\leq 2\eta_{\revise{k}} \delta^2 \sum_{e\not\in S_m} \pi_{em} \|\mu_e - w^*_m\|_2^2 - 4\eta_{\revise{k}} \delta^2 \sum_{e\in S_m}\pi_{em} \|\theta_{em}^{\revise{(k)}} - \mu_e\|_2 \|\mu_e-w^*_m\|_2\\
    &\leq 2\eta_{\revise{k}} \delta^2 E(1-p) + 4\eta_{\revise{k}} \delta^2 \gamma_m r
\end{align}

we have
\begin{equation}
    \mathbb E\|\hat w_m^{\revise{(k)}} - w^*_m - \eta_{\revise{k}} \hat g_{\revise{k}}\|^2 \leq (2\eta_{\revise{k}} \delta^2 \gamma_m p)\mathbb E\|\hat w_m^{\revise{(k)}} - w^*_m\|^2 + 16\eta_{\revise{k}}^2 E(\revise{K}-1)^2 \delta^4 + 2\eta_{\revise{k}}\delta^2 E(1-p) + 4\eta_{\revise{k}} \delta^2 \gamma_m r
\end{equation}

Notice that
\begin{align}
    \mathbb E \|\hat g_{\revise{k}} - g_{\revise{k}}\|^2 &=\mathbb E \sum_{e} \frac{4}{n_e^2}\pi_{em} \|(X_e^T X_e - \mathbb E (X_e^T X_e))(\theta_{em}^{\revise{(k)}} - \mu_e)\|^2 + \mathbb E\sum_{e}\frac{4}{n_e^2} \sum_{e}\pi_{em} \|X_e^T \epsilon_e\|^2\nonumber\\
    &= E\frac{O(dn_e)}{n_e^2} \delta^4 +E \frac{O(d n_e)}{n_e^2}\delta^2 \sigma^2
\end{align}

so
\begin{equation}
    \mathbb E\|\hat w_m^{\revise{(k+1)}} - w^*_m\|_2^2 \leq (1 - 2\eta_{\revise{k}} \gamma_m p\delta^2)\mathbb E\|\hat w_m^{\revise{(k)}} - w^*_m\|_2^2 + \eta_{\revise{k}} A_1 + \eta_{\revise{k}}^2 A_2 
\end{equation}
where 
\begin{equation}
    A_1 = 4\delta^2 \gamma_m r + 2\delta^2 E(1-p)
\end{equation}
and
\begin{equation}
    A_2 =16E(\revise{K}-1)^2 \delta^4 + O(\frac{d}{n_e}) E (\delta^4 + \delta^2 \sigma^2).
\end{equation}

\end{proof}

\subsection{Convergence of Models with Smooth and Strongly Convex Losses (\cref{theorem:convex_converge})}
Here we present the detailed proof for \cref{theorem:convex_converge}.

\subsubsection{Key Lemmas}
We first state two lemmas for E-step updates and M-step updates, respectively. The proofs of both lemmas are deferred to the \cref{appendixA2.2}
\begin{lemma}
\label{lemma:3}
Suppose the loss function $\mathcal L_{P_t}(\theta)$ is $L$-smooth and $\mu$-strongly convex for any cluster $m$. If $\|w_m^{\revise{(t)}} - w_m^*\|\leq \frac{\sqrt{\mu}R}{\sqrt{\mu}+\sqrt{L}} - r - \Delta$ for some $\Delta>0$, then E-step updates as
\begin{equation}
    \pi_{em}^{\revise{(t)}} \geq \frac{\pi_{em}^{\revise{(t)}}}{\pi_{em}^{\revise{(t)}} + (1-\pi_{em}^{\revise{(t)}}) \exp(-\mu R\Delta)}.
\end{equation}
\end{lemma}

For M-steps, the local agents are initialized with $\theta^{(0)}_{em} = w_m^{\revise{(t)}}$. Then for $\revise{k}=1,\dots, \revise{K}-1$, each agent use local SGD to update its personal model:
\begin{equation}
    \theta_{em}^{\revise{(k+1)}} = \theta_{em} - \eta_{\revise{k}} g_{em}(\theta_{em}) = \theta_{em}^{\revise{(k)}} - \eta_{\revise{k}} \nabla\sum_{i=1}^{n_e} \ell(h_{\theta_{em}}(x_e^{(i)}),y_e^{(i)}).
\end{equation}

To analyze the aggregated model \cref{M:agg}, we define a sequence of virtual aggregated models $\hat w_m^{\revise{(k)}}$.
\begin{equation}
\label{eq:50}
    \hat w_m^{\revise{(k)}} = \sum_{e=1}^E\frac{\pi_{em}\theta_{em}^{\revise{(k)}}}{\sum_{e^\prime = 1}^E \pi_{e^\prime m}}.
\end{equation}

\begin{lemma}
\label{lemma:4}
Suppose for any agent $e\in S_m$, its soft clustering label $\pi_{em}^{\revise{(t+1)}} \geq p$. Then one step local SGD updates $\hat w_m^{\revise{(k)}}$ by \cref{eq:lemma4}, if the learning rate $\eta_{\revise{k}} \leq \frac{1}{2(\mu + L)}$.
\begin{equation}
\label{eq:lemma4}
    \mathbb E\|\hat w_m^{\revise{(k+1)}} - w_m^*\|_2^2 \leq (1 -  \eta_{\revise{k}} A_0) \mathbb E\|\hat w_m^{\revise{(k)}} - w_m^*\|_2^2 + \eta_{\revise{k}} A_1 + \eta_{\revise{k}}^2 A_2.
\end{equation}
where 
\begin{equation}
    A_0 = \frac{2\gamma_m p\mu L}{\mu + L}
\end{equation}
\begin{equation}
    A_1 = 2\gamma_m L r \sqrt{\frac{2G}{\mu}} + \frac{G(1-p)E}{\mu} (4L+ \frac{6}{\mu + L}) +O(r^2).
\end{equation}
\begin{equation}
    A_2 =\frac{4E(\revise{K}-1)^2 GL^2}{\mu} +  \frac{E\sigma^2}{n_e}.
\end{equation}
\end{lemma}
\textbf{Remark.} Using this recursive relation, if the learning rate $\eta_{\revise{k}}$ is fixed, the sequence $\hat w_m^{\revise{(k+1)}}$ has a convergence rate of
\begin{equation}
    \mathbb E\|\hat w_m^{\revise{(k)}}-w_m^*\|^2\leq (1 - \eta A_0)^{\revise{k}} \mathbb E\|\hat w_m^{(0)} - w_m^*\|^2 + \eta \revise{k} (A_1 + \eta A_2).
\end{equation}

\subsubsection{Completing the Proof of \cref{theorem:convex_converge}}

\begingroup
\def\thetheorem{\ref{theorem:convex_converge}}
\begin{theorem}

Suppose loss functions have bounded variance for gradients on local datasets, i.e., ${\footnotesize\mathbb E_{(x,y)\sim \mathcal D_e} [\|\nabla\ell (x,y;\theta) - \nabla \mathcal L_e(\theta)\|_2^2]\leq \sigma^2}$. Assume population losses are bounded, i.e., $\mathcal L_e\in G, \forall e\in [E]$.
With initialization from assumptions 3 and 4, if each agent chooses learning rate $\eta \leq \min(\frac{1}{2(\mu+L)}, \frac{\beta}{\sqrt{\revise{T}}})$, the weights $(\Pi, W)$ converges by
\begin{align}
    & \pi_{em}^{\revise{(T)}} \geq \frac{1}{1 + (M-1)\exp(-\mu R\Delta_0 \revise{T})},\ \forall  e\in S_m\label{eq:pi_evolve_}\\
    & \mathbb E\|w_m^{\revise{(T)}} - w_m^*\|_2^2 \leq (1 - \eta A)^{KT} (\|w_m^{(0)} - w_m^*\|_2^2 + B) + O(\revise{K}r) + \frac{ME\beta O(\revise{K}^3,\frac{\sigma^2}{n_e})}{\sqrt{\revise{T}}}
    \label{eq:w_evolve_}
\end{align}
where $\revise{T}$ is the total number of communication rounds; $\revise{K}$ is the number of iterations each round; $\gamma_m = |S_m|$ is the number of agents in the $m$-th cluster, and
\small
\begin{equation}
    A = \frac{2\gamma_m}{M}\frac{\mu L}{\mu + L}, 
    B = \frac{GMTE(\frac{4L}{\mu}+\frac{6}{\mu(\mu+L)})}{(1 - \eta A)^{\revise{K}} - \exp(-\mu R\Delta_0)}.
\end{equation}
\normalsize
\end{theorem}
\addtocounter{theorem}{-1}
\endgroup

\begin{proof}
The proof is quite similar to Theorem 1 for linear models: we follow an induction proof using lemmas 3 and 4. Suppose \cref{eq:pi_evolve_} hold for step $\revise{t}$. And suppose
\small
\begin{equation}
    \mathbb E\|w_m^{\revise{(t)}} - w_m^*\|_2^2 \leq (1 - \eta A)^{\revise{Kt}} (\|w_m^{(0)} - w_m^*\|_2^2) + B((1-\eta A)^{\revise{Kt}} - \exp(-\mu R \Delta_0 \revise{t})) + \frac{\eta C }{1 - (1 - \eta A)^{\revise{K}}}. \label{eq:91}
\end{equation}
\normalsize
where
\begin{equation}
    C = \frac{4\eta EG\revise{K}^3L^2}{\mu} + (2\gamma_m Lr \sqrt{\frac{2G}{\mu}} + O(r^2)) +  \eta \frac{E\revise{K}\sigma^2}{n_e}.
\end{equation}

Then for any $t\in S_m$,
\begin{align}
    \pi_{em}^{\revise{(t+1)}} &\geq \frac{\pi_{em}^{\revise{(t)}}}{\pi_{em}^{\revise{(t)}} + (1 - \pi_{em}^{\revise{(t)}})\exp(-\mu R \Delta_{\revise{t}})}\\
    &\geq \frac{1}{1 + (M-1) \exp(-\mu R \Delta_0 \revise{t}) \exp(-\mu R \Delta_{\revise{t}})}\\
    & \geq \frac{1}{1 + (M-1) \exp(-\mu R \Delta_0 (\revise{t}+1))}
\end{align}

We recall the virtual sequence $\hat w_m^{\revise{(k)}}$ defined in \cref{eq:50}. Models are synchronized after $\revise{K}$ rounds of local iterations, so $w_m^{\revise{(t+1)}} = \hat w_m^{\revise{(K)}}$. Thus, according to \cref{lemma:4},
{\small
\begin{align}
  &\mathbb E\| w_m^{\revise{(t+1)}} - w_m^*\|_2^2 = \mathbb E\|\hat w_m^{\revise{(K)}} - w_m^*\|_2^2\\
  &\leq (1 - \eta A_0)^{\revise{K}} \mathbb E\|w_m^{\revise{(t)}} - w_m^*\|_2^2 + \eta \revise{K} (A_1 + \eta A_2)\\
  &\leq (1 - \eta A_0)^{\revise{K}} \Big((1 - \eta A)^{\revise{Kt}} (\mathbb E \|w_m^{(0)} - w_m^*\|^2) + B((1 - \eta A)^{\revise{Kt}} - \exp(-\mu R\Delta_0 \revise{t})) + \frac{\eta C} {1 - (1 - \eta A)^{\revise{K}}} \Big) + \eta \revise{K}(A_1 + \eta A_2)\\
  &\leq (1 - \eta A)^{(\revise{t}+1)\revise{K}} \mathbb E \|w_m^{(0)} - w_m^*\|^2 + \underbrace{ (1 - \eta A)^{\revise{K}} B\big((1 - \eta A)^{\revise{Kt}} - \exp(-\mu R\Delta_0 \revise{t})\big) + \eta \frac{G\revise{K}(1 - p)E}{\mu}(4L + \frac{6}{\mu + L})}_{F_1}\nonumber \\ 
  & \hspace{3em} + \underbrace{(1 - \eta A)^{\revise{K}} \frac{\eta C}{1 - (1 - \eta A)^{\revise{K}}} + \eta \revise{K} (2\gamma_m Lr \sqrt{\frac{2G}{\mu}} + O(r^2)) + \eta^2 \revise{K} A_2}_{F_2}.
\end{align}
}
For $F_1$, we use the fact that 
$$\pi_{em}^{\revise{(t+1)}} \geq \frac{1}{1 + (M-1)\exp-(\mu R\Delta_0(\revise{t}+1))} \geq 1 - (M-1)\exp (-\mu R\Delta(\revise{t}+1)),$$
so
\begin{align}
    F_1 &\leq   (1 - \eta A)^{\revise{K}} B\big((1 - \eta A)^{\revise{Kt}} - \exp(-\mu R\Delta_0 \revise{t})\big) + \eta \frac{G(M-1)\exp(-\mu R \Delta_0 \revise{t})}{\mu}(4 L + \frac{6}{\mu + L})\\
    &= B \Big((1 - \eta A)^{(\revise{t}+1)\revise{K}} - \exp(- \mu R \Delta_0 \revise{t})\Big)
\end{align}

For $F_2$, we have
\begin{align}
    F_2 &\leq (1 - \eta A)^{\revise{K}} \frac{\eta C}{1 - (1 - \eta A)^{\revise{K}}} +  \eta \revise{K}(2\gamma_m Lr \sqrt{\frac{2G}{\mu}} + O(r^2)) + \frac{4EGL^2\eta^2 \revise{K}^3}{\mu} +  \frac{\eta^2 \revise{K}E \sigma^2}{n_e}\\
    &\leq \frac{\eta C}{1 - (1 - \eta A)^{\revise{K}}}.
\end{align}
Combining $F_1$ and $F_2$ finishes the induction proof. Moreover, since $T\geq 1$, we have
\begin{equation}
    \frac{\eta C}{1 - (1 - \eta A)^{\revise{K}}} \leq \frac{C}{A} = O(\revise{K}r) + \frac{ME\beta}{\sqrt{\revise{T}}} O(\revise{K}^3, \frac{\sigma^2}{n_e}).\label{eq:104}
\end{equation}
Combining \cref{eq:91} and \cref{eq:104} completes our proof.
\end{proof}

\subsubsection{Deferred Proofs of Key Lemmas}
\label{appendixA2.2}

\textbf{Lemma 3.}
\begin{proof}
According to \cref{algorithm1},
\begin{align}
    \pi_{em}^{\revise{(t+1)}} &= \frac{\pi_{em}^{\revise{(t)}}}{\pi_{em}^{\revise{(t)}} + \sum_{m^\prime\neq m} \pi_{em^{\prime}}^{\revise{(t)}} \exp\Big(\mathbb E \ell(x,y;w_m^{\revise{(t)}}) - \mathbb E \ell(x,y;w_{m^\prime}^{\revise{(t)}}) \Big)}\\
    &\geq \frac{\pi_{em}^{\revise{(t)}}}{\pi_{em}^{\revise{(t)}} + (1-\pi_{em}^{\revise{(t)}}) \exp\Big(\max_{m^\prime \neq m} (\mathcal L_{P_e}(w_m^{\revise{(t)}}) - \mathcal L_{P_e}(w_{m^\prime}^{\revise{(t)}}))\Big)}\label{eq:82}
\end{align}

Since $\mathcal L_{P_e}$ is $L$-smooth and $\mu$-strongly convex,
\begin{align}
    \mathcal L_{P_e}(w_m^{\revise{(t)}}) - \mathcal L_{P_e}(w_{m^\prime}^{\revise{(t)}}) &\leq \frac{L}{2}\|w_m^{\revise{(t)}} - \theta_t^*\|^2 - \frac{\mu}{2}\|w_{m^\prime}^{\revise{(t)}} - \theta_t^*\|^2\nonumber\\
    &\leq \frac{L}{2} (\frac{\sqrt{\mu}R}{\sqrt{\mu} + \sqrt{L}} - \Delta)^2 - \frac{\mu}{2} (\frac{\sqrt{L}R}{\sqrt{\mu} + \sqrt{L}} + \Delta)^2\nonumber\\
    &\leq -\sqrt{\mu L}R\Delta + \frac{L-\mu}{2}\Delta^2 \leq -\mu R\Delta.\label{eq:83}
\end{align}
Combining \cref{eq:82} and \cref{eq:83} completes our proof.
\end{proof}

\textbf{Lemma 4.}
\begin{proof}

We define $g_{m}^{\revise{(k)}} = \sum_{e} \pi_{em} \frac{1}{n_e}\sum_{i=1}^{n_e}\nabla \ell(h_{\theta_{em}}(x_e^{(i)}), y_e^{(i)})$ and $\hat g_m^{\revise{(k)}} = \sum_{e} \pi_{em} \nabla \mathcal L(\theta_{em}^{\revise{(k)}})$.
\begin{align}
    \mathbb E\|\hat w_m^{\revise{(k+1)}} - w_m^* \|^2 &= \mathbb E\|\hat w_m^{\revise{(k)}} - w_m^* - \eta_{\revise{k}} g_{m}^{\revise{(k)}}\|^2 \\
    &= \mathbb E\|\hat w_m^{\revise{(k)}} - w_m^* - \eta_{\revise{k}} \hat g_m^{\revise{(k)}}\|^2 + \eta_{\revise{k}}^2 \mathbb E\|g_m^{\revise{(k)}} - \hat g_m^{\revise{(k)}}\|^2 \nonumber \\
    &\hspace{8em}+ 2\eta_{\revise{k}}\mathbb E \langle w_m^{\revise{(k)}} - w_m^* - \eta_{\revise{k}} \hat g_m^{\revise{(k)}}, \hat g_m^{\revise{(k)}} - g_m^{\revise{(k)}}\rangle\\
    &=  \mathbb E\|\hat w_m^{\revise{(k)}} - w_m^* - \eta_{\revise{k}} \hat g_m^{\revise{(k)}}\|^2 + \eta_{\revise{k}}^2 \mathbb E\|g_m^{\revise{(k)}} - \hat g_m^{\revise{(k)}}\|^2.
\end{align}

The first term can be decomposed into
\begin{equation}
    \|\hat w_m^{\revise{(k)}} - w_m^* - \eta_{\revise{k}} \hat g_m^{\revise{(k)}}\|^2 = \|\hat w_m^{\revise{(k)}} - w_m^*\|^2 + \eta_{\revise{k}}^2\|\hat g_m^{\revise{(k)}}\|^2 - 2\eta_{\revise{k}} \langle \hat w_m^{\revise{(k)}} - w_m^*,\hat g_m^{\revise{(k)}}\rangle.
\end{equation}

Note that
\begin{align}
    &\|\hat g_m^{\revise{(k)}}\|^2  \leq \sum_{e=1}^E\pi_{em}\|\nabla \mathcal L_e(\theta_{em}^{\revise{(k)}})\|^2.\\
    &-\langle \hat w_m^{\revise{(k)}} - w_m^*,\hat g_m^{\revise{(k)}}\rangle =  -\sum_{e=1}^E \pi_{em}\langle \hat w_m^{\revise{(k)}} - \theta_{em}^{\revise{(k)}}, \nabla \mathcal L_e(\theta_{em}^{\revise{(k)}}) \rangle - \sum_{e=1}^E \pi_{em} \langle \theta_{em}^{\revise{(k)}} - w_m^*, \nabla \mathcal L_e(\theta_{em}^{\revise{(k)}})\rangle.\label{eq:73}
\end{align}

We further decompose the two terms in \cref{eq:73} by
\begin{equation}
    -2\langle \hat w_m^{\revise{(k)}} - \theta_{em}^{\revise{(k)}}, \nabla \mathcal L_e(\theta_{em}^{\revise{(k)}}) \rangle \leq \frac{1}{\eta_{\revise{k}}} \|\hat w_m^{\revise{(k)}} - \theta_{em}^{\revise{(k)}}\|^2 + \eta_{\revise{k}} \|\nabla \mathcal L_e(\theta_{em}^{\revise{(k)}})\|^2.
\end{equation}
and
\begin{align}
    \langle \theta_{em}^{\revise{(k)}} - w_m^*, \nabla \mathcal L_e(\theta_{em}^{\revise{(k)}})\rangle &\geq \langle \theta_{em}^{\revise{(k)}} - w_m^*, \nabla \mathcal L_e(\theta_{em}^{\revise{(k)}}) - \nabla \mathcal L_e(w_m^*)\rangle - \|\nabla \mathcal L_e(w_m^*)\|_2 \|\theta_{em}^{\revise{(k)}} - w_m^*\|_2.\\
    &\geq \frac{\mu L}{\mu + L} \|\theta_{em}^{\revise{(k)}} - w_m^*\|^2 + \frac{1}{\mu + L}\|\nabla \mathcal L_e(\theta_{em}^{\revise{(k)}} - \nabla \mathcal L_e(w_m^*))\|^2 - \|\nabla \mathcal L_e(w_m^*)\|_2 \|\theta_{em}^{\revise{(k)}} - w_m^*\|_2.
\end{align}

Therefore,
\begin{align}
    \mathbb E\|\hat w_m^{\revise{(k+1)}} - w_m^*\|^2 &\leq 
    \underbrace{\mathbb E\|\hat w_m^{\revise{(k)}} - w_m^*\|^2 - 2\eta_{\revise{k}} \frac{\mu L}{\mu + L}\sum_{e}\pi_{em}\mathbb E\|\theta_{em}^{\revise{(k)}} - w_m^*\|^2}_{E_1}+\underbrace{\sum_{e}\pi_{em}\mathbb E\|\hat w_m^{\revise{(k)}} - \theta_{em}^{\revise{(k)}}\|^2}_{E_2}  \nonumber\\
    &\hspace{3em}+\underbrace{\Big(2 \eta_{\revise{k}}^2 \sum_{e}\pi_{em}\mathbb E \|\nabla \mathcal L_e(\theta_{em}^{\revise{(k)}})\|^2 - 2\eta_{\revise{k}} \frac{1}{\mu + L}\sum_{e}\pi_{em}\mathbb E\|\nabla\mathcal L_e(\theta_{em}^{\revise{(k)}})- \nabla \mathcal L_e(w_m^*)\|^2\Big)}_{E_3}
    \nonumber\\
    &\hspace{3em}+ \underbrace{2\eta_{\revise{k}} \mathbb E \sum_{e}\pi_{em}\|\theta_{em}^{\revise{(k)}} - w_m^*\|_2 \cdot \|\nabla \mathcal L_e(w_m^*)\|_2}_{E_4} + \underbrace{\eta_{\revise{k}}^2 \mathbb E\|g_m^{\revise{(k)}} - \hat g_m^{\revise{(k)}}\|^2}_{E_5}.
\end{align}
\end{proof}

\begin{align}
    E_1 &= \mathbb E \|\hat w_m^{\revise{(k)}} - w_m^*\|^2 - 2\eta_{\revise{k}} \frac{\mu L}{\mu + L}\mathbb E\Big(\sum_{e} \pi_{em}\|\hat w_m^{\revise{(k)}} - w_m^*\|^2 + \sum_{e}\pi_{em}\|\hat w_m^{\revise{(k)}} - \theta_{em}^{\revise{(k)}}\|^2 \Big)\nonumber \\
    &\leq (1 - \frac{2\eta_{\revise{k}} \mu L p \gamma_m}{\mu + L})\mathbb E\|w_m^{\revise{(k)}} - w_m^*\|^2 + E_2.\label{E1}
\end{align}

\begin{align}
    E_2 &= \mathbb E \sum_{e}\pi_{em} \|\hat w_m^{\revise{(k)}} - \theta_{em}^{\revise{(k)}}\|^2\nonumber \\
    &= \mathbb E \sum_{e}\pi_{em}\|(w_m^{(0)} - \theta_{em}^{\revise{(k)}}) + (\theta_{em}^{\revise{(k)}} - w_m^{\revise{(k)}})\|^2\nonumber \\
    &\leq \mathbb E \sum_{e}\pi_{em}\|(w_m^{(0)} - \theta_{em}^{\revise{(k)}})\|^2\nonumber \\
    &\leq \sum_{e}\pi_{em}(\revise{K}-1)\mathbb E \sum_{\revise{k}^\prime=0}^{\revise{k}-1} {\eta_{\revise{k}^\prime}}^2 \|g_{em}(\theta_{em}^{(\revise{k}^\prime)})\|^2\nonumber \\
    &\leq \frac{2\eta_{\revise{k}}^2 E(\revise{K}-1)^2 G^2L^2}{\mu}.\label{E2}
\end{align}

\begin{align}
    E_3 &= 2\mathbb E\sum_{e}\pi_{em}\Big((\eta_{\revise{k}}^2 - \frac{\eta_{\revise{k}}}{\mu + L}) \|\nabla \mathcal L_e(\theta_{em}^{\revise{(k)}})\|^2 + \frac{2\eta_{\revise{k}}}{\mu + L}\langle \nabla \mathcal L_e(\theta_{em}^{\revise{(k)}}), \nabla \mathcal L_e(w_m^*)\rangle - \eta_{\revise{k}} \frac{\|\nabla \mathcal L_e(w_m^*)\|^2}{\mu + L}\Big)\nonumber \\
    &\leq 4\eta_{\revise{k}}\mathbb E \sum_{e}\pi_{em}\Big( -\frac{1}{2(\mu + L)} \|\nabla \mathcal L_e(\theta_{em}^{\revise{(k)}})\|^2 + \frac{1}{\mu + L}\langle \nabla \mathcal L_e(\theta_{em}^{\revise{(k)}}), \nabla \mathcal L_e(w_m^*)\rangle -\frac{\|\nabla \mathcal L_e(w_m^*)\|^2}{\mu + L}\Big)\nonumber\\
    &\leq 6\eta_{\revise{k}}\sum_{e}\pi_{em} \frac{\|\nabla \mathcal L_e(w_m^*)\|^2}{\mu + L}\nonumber \\
    &\leq 6\eta_{\revise{k}} \sum_{e\in S_m} \pi_{em} \frac{L^2 r^2}{\mu + L} + 6\eta_{\revise{k}} \sum_{e\not\in S_m} \pi_{em} \frac{2G}{\mu(\mu + L)}\nonumber \\
    &\leq \eta_{\revise{k}} O(r^2) + 6\eta_{\revise{k}} \frac{G(1-p)E}{\mu(\mu + L)}. \label{E3}
\end{align}

\begin{align}
    E_4 &= 2\eta_{\revise{k}}\mathbb E\sum_{e\in S_m} \pi_{em} \|\theta_{em}^{\revise{(k)}} - w_m^*\|_2 \cdot \|\nabla \mathcal L_e(w_m^*)\|_2+ 2\eta_{\revise{k}}\mathbb E\sum_{e\not\in S_m} \pi_{em} \|\theta_{em}^{\revise{(k)}} - w_m^*\|_2 \cdot \|\nabla \mathcal L_e(w_m^*)\|_2 \nonumber \\
    & \leq 2\eta_{\revise{k}} \gamma_m Lr \sqrt{\frac{2G}{\mu}} + 2\eta_{\revise{k}} (1-p) EL\cdot \frac{2G}{\mu}.\label{E4}
\end{align}

\begin{align}
    E_5 &=  \eta_{\revise{k}}^2 \mathbb E \|g_m^{\revise{(k)}} - \hat g_m^{\revise{(k)}}\|^2 \nonumber \\
    &\leq \eta_{\revise{k}}^2 \mathbb E \Big\|\sum_e \pi_{em}\Big( \frac{1}{n_e} \sum_{i=1}^{n_e} \nabla \ell(h_{\theta_{em}}(x_e^{(i)}), y_e^{(i)}) - \mathcal L(\theta_{em}^{\revise{(k)}})\Big)\Big\|^2\nonumber \\
    &\leq \eta_{\revise{k}}^2 E \frac{\sigma^2}{n_e}.\label{E5}
\end{align}

Combining \cref{E1} to \cref{E5} yields the conclusion of \cref{lemma:4}.

\section{Fairness Analysis}
\label{appendixB}
\subsection{Proof of \cref{fairness_linear}}
\label{appendix:B1}
\begin{proof}
Let the first cluster $m_1$ contain agents $\mu_1,\dots, \mu_{E-1}$, while the second cluster contains only the outlier $\mu_E$. Then, for $e= 1,\dots, E-1$,
\begin{equation}
    \mathcal E_{e}(w_{m_1}) = \delta^2 \norm{\mu_e - \frac{\sum_{e^\prime=1}^{E-1} \mu_{e^\prime}}{E-1}}^2 \leq \delta^2 r^2
\end{equation}
And for the outlier agent, the expected output is just the optimal solution, so
\begin{equation}
    \mathcal E_{E}(w_{m_2}) = 0
\end{equation}

As a result, the fairness of this algorithm is bounded by
\begin{equation}
    \faa_{focus}(P) = \max_{i,j\in [E]} |\mathcal E_i(\Pi, W) - \mathcal E_j(\Pi, W)| \leq \delta^2 r^2.
\end{equation}

On the other hand, the expected final weights of of FedAvg algorithm is $w_{avg} = \bar \mu = \frac{\sum_{e=1}^E\mu_e}{E}$, so the expected loss for agent $e$ shall be
\begin{equation}
    \mathbb E_{(x,y)\sim \mathcal P_e}(\ell_{\hat\theta}(x)) = \mathbb E_{x\sim \mathcal N(0,\delta^2 I_d), \epsilon \sim \mathcal N(0,\sigma_e^2)}[(\mu_i^T x+ \epsilon - \bar \mu^T x)^2] = \sigma_e^2 + \delta^2 \norm{\mu_e - \bar\mu}^2
\end{equation}
The infimum risk for agent $t_1$ is $\sigma_1^2$, and after subtracting it from the expected loss, we have
\begin{align}
        \mathcal E_{1}(w_{avg}) &= \delta^2 \norm{\mu_1 - \bar \mu}^2\\
        &= \delta^2 \|\mu_1 - \frac{\sum_{e=1}^{E-1}\mu_1}{E} - \frac{\mu_E}{E}\|^2 \\ 
        &\leq \delta^2\Big(r\cdot \frac{E-1}{E} + \frac{\norm{\mu_1 - \mu_E}}{E}\Big)^2 \\
        &\leq \delta^2(r\cdot \frac{E-1}{E} + \frac{R+r}{E})^2 = \delta^2 (r + \frac{R}{E})^2
\end{align}
However for the outlier agent,
\begin{align}
    \mathcal E_{E}(w_{avg}) &= \delta^2 \|\mu_E - \bar\mu\|^2 \\
    &= \delta^2 \norm{\frac{E-1}{E} \mu_E - \frac{\sum_{e=1}^{E-1}\mu_E}{E}}^2\\
    &\geq \Big(\frac{E-1}{E}\Big)^2 \delta^2 R^2
\end{align}
Hence,
\begin{equation}
    \faa_{avg}(P) \geq \mathcal E_{E}(w_{avg}) - \mathcal E_{1}(w_{avg}) = \delta^2\Big(\frac{R^2(E-2) - 2Rr}{E} + r^2\Big)
\end{equation}
\end{proof}

\textbf{Remark.}
When there are $E_k > 1$ outliers, we can similarly derive FAA for FedAvg algorithm:
\begin{equation}
    \mathcal E_1(w_{avg}) \leq \delta^2 (r+\frac{E_k R}{E})^2
\end{equation}
\begin{equation}
    \mathcal E_{E}(w_{avg})\geq \delta^2 (\frac{E-E_k}{E} R - \frac{E_k}{E}r)^2
\end{equation}
so as long as $E_k < \frac{E}{2}$,
\begin{equation}
    \faa_{avg} \geq \mathcal E_{E}(w_{avg}) - \mathcal E_1(w_{avg}) = \Omega(\delta^2 R^2)
\end{equation}
The FOCUS algorithm produces a result with
\begin{equation}
    \mathcal E_{1} (w_{m_1}) \leq \delta^2 r^2
\end{equation}
\begin{equation}
    \mathcal E_E (w_{m_2})\leq \delta^2 r^2
\end{equation}
Hence we still have
\begin{equation}
    \faa_{focus} \leq \delta^2 r^2.
\end{equation}

\subsection{Proof of \cref{fairness_convex}}
\label{appendix:B2}
\begin{proof}
Note that the local population loss for agent $i$ with weights $\theta$ is
\begin{equation}
    \mathcal L_{i}(\theta) = \int p_i(x,y) \ell(f_{\theta}(x), y)\mathrm dx\mathrm dy.
\end{equation}
Thus,
\begin{align}
    |\mathcal L_{i}(\theta_i^*) - \mathcal L_{j}(\theta_i^*)| &= \int |p_i(x,y) - p_j(x,y)| \cdot \ell(f_{\theta_i^*}(x), y)\mathrm dx\mathrm dy\\
    &\leq G \cdot \int |p_i(x,y) - p_j(x,y)|\mathrm dx \mathrm dy \leq Gr.
\end{align}

Hence,
\begin{equation}
    \mathcal L_{i}(\theta_j^*) \leq \mathcal L_{j}(\theta_j^*) + Gr \leq \mathcal L_{j}(\theta_i^*) +Gr \leq  \mathcal L_{i}(\theta_i^*) + 2Gr.
\end{equation}
For the cluster that combines agents $\{1,\dots, E-1\}$ together, the weight converges to $\bar\theta^\prime~=~\frac{1}{E-1}~\sum_{i=1}^{E-1}~\theta_i^*$. Then $\forall i = 1,\dots, E-1$, the population loss for the ensemble prediction
\begin{align}
    \mathcal L_{i}(\theta, \Pi) &= \mathcal L_{i}\Big(\frac{\sum_{j=1}^{E-1}\theta_{j}^*}{E-1}\Big)\\
    &\leq \frac{1}{T-1} \sum_{j=1}^{T-1} \mathcal L_{i}(\theta_j^*)\\
    &\leq \mathcal L_{i}(\theta_i^*) + \frac{2Gr}{E-1}.
\end{align}

Therefore, for any $i = 1,\dots, T-1$,
\begin{equation}
    \mathcal E_{i}(\theta, \Pi) \leq \frac{2Gr}{E-1}.
\end{equation}
Since $\mathcal E_{T}(\theta, \Pi) = 0$, 
\begin{equation}
    \faa_{focus}(W,\Pi) \leq \frac{2Gr}{E-1}
\end{equation}

Now we prove the second part of \cref{fairness_convex} for the fairness of Fedavg algorithm. For simplicity, we define $B = \frac{2Gr}{E-1}$ in this proof. Also, we denote the mean of all optimal weight $\bar \theta = \frac{\sum_{i=1}^E \theta_i^*}{E}$ and $\bar \theta^\prime = \frac{\sum_{i=1}^{E-1} \theta_i^*}{E-1}$. 

Remember that we assume loss functions to be L-smooth, so
\begin{equation}
    \mathcal L_{E}(\theta_i^*)\leq \mathcal L_{E}(\bar\theta^\prime) + \langle \nabla \mathcal L_{E}(\bar\theta^\prime), \theta_i^* - \bar\theta^\prime \rangle + \frac{L}{2}\|\bar\theta^\prime - \theta_i\|^2.
\end{equation}

Taking summation over $i = 1,\dots, E-1$, we get
\begin{align}
    \mathcal L_{E}(\bar\theta^\prime) &\geq \frac{1}{E-1}\Big(\sum_{i=1}^{E-1} \mathcal L_{E}(\theta_i^*) - \langle \nabla \mathcal L_{E}(\bar\theta^\prime), \sum_{i=1}^{E-1}(\theta_i - \bar\theta^\prime)\rangle - \frac{L}{2}\sum_{i=1}^{E-1}\|\bar\theta^\prime - \theta_i\|^2\Big)\\
    &= \frac{1}{E-1}\Big(\sum_{i=1}^{E-1} \mathcal L_{E}(\theta_i^*) - \frac{L}{2}\sum_{i=1}^{E-1}\|\bar\theta^\prime - \theta_i\|^2\Big)\\
    &\geq \mathcal L_{E}(\theta_E^*) + R - \frac{LB}{\mu}. \label{eq:mean_prime}
\end{align}
The last inequality uses the $\mu$-strongly convex condition that implies
\begin{equation}
    B \geq \mathcal L_{i}(\bar\theta^\prime) - \mathcal L_{i}(\theta_i^*) \geq \frac{\mu}{2} \|\bar\theta^\prime - \theta_i\|^2.
\end{equation}

By $L$-smoothness, we have
\begin{align}
    & \mathcal L_{E}(\bar\theta^\prime) \leq \mathcal L_{E}(\bar\theta) + \langle \nabla \mathcal L_{E}(\bar\theta), \bar\theta^\prime - \bar\theta \rangle + \frac{L}{2} \|\bar\theta^\prime - \bar\theta\|^2.\\
    &\mathcal L_{E}(\theta_E^*) \leq \mathcal L_{E}(\bar\theta) + \langle \nabla \mathcal L_{E}(\bar\theta), \theta_E^* - \bar\theta\rangle + \frac{L}{2} \|\theta_E^* - \bar\theta\|^2.
\end{align}

Note that $\bar\theta = \frac{\bar\theta^\prime + (E-1)\theta_E^*}{E}$, we take a weighted sum over the above two inequalities to cancel the dot product terms out. We thus derive
\begin{align}
    \mathcal L_{E}(\bar\theta) &\geq \frac{(E-1)\mathcal L_{E}(\bar\theta^\prime) + \mathcal L_{E}(\theta_E^*) - \frac{L}{2} (E-1)\|\bar\theta^\prime - \bar\theta\|^2 - \frac{L}{2} \|\theta_E^* - \bar\theta\|^2}{E}\\
    &= \frac{E-1}{E} \Big(R -\frac{LB}{\mu} -  \frac{L\|\theta_E^* - \bar\theta^\prime\|^2}{2E}\Big) + \mathcal L_{E}(\theta_E^*).
\end{align}

Note that $\mathcal L_{E}(\cdot)$ is $\mu$-strongly convex, which means
\begin{equation}
    R- \frac{LB}{\mu} \geq \mathcal L_{E}(\bar\theta^\prime) - \mathcal L_{E}(\theta_E^*) \geq \frac{\mu}{2}\|\theta_E^* - \bar\theta^\prime\|^2.
\end{equation}
so
\begin{equation}
    \mathcal L_{E}(\bar\theta) \geq (1-\frac{L}{\mu E})\cdot \frac{E-1}{E} (R - \frac{LB}{\mu}) + \mathcal L_{E}(\theta_E^*).
\end{equation}
And
\begin{equation}
    \mathcal E_{E}(\bar\theta) \geq (1-\frac{L}{\mu E})\cdot \frac{E-1}{E} (R - \frac{LB}{\mu}).
\end{equation}

On the other hand, for agent $i = 1,\dots, E-1$ we know
\begin{equation}
    \mathcal L_{i}(\bar\theta) \leq \mathcal L_{i}(\bar\theta^\prime) + \langle \nabla \mathcal L_{i}(\bar\theta^\prime), \bar\theta - \bar\theta^\prime\rangle + \frac{L}{2}\|\bar\theta - \bar\theta^\prime\|^2.
\end{equation}

By $L$ smoothness,
\begin{equation}
    \|\nabla \mathcal L_{i}(\bar\theta^\prime)\|_2 \leq L\|\bar\theta^\prime - \theta_i^*\| \leq L\sqrt{\frac{2B}{\mu}}.
\end{equation}

So
\begin{equation}
    \mathcal L_{i}(\bar\theta) \leq \mathcal L_{i}(\theta_i^*) + B + L \sqrt{\frac{2B}{\mu}} \sqrt{\frac{2(R - \frac{LB}{\mu})}{\mu}}\frac{1}{E} + \frac{L (R - \frac{LB}{\mu})}{\mu E^2}
\end{equation}
\begin{equation}
    \mathcal E_{i}(\bar\theta) \leq B + \frac{2L}{\mu E} \sqrt{B(R - \frac{LB}{\mu})} + \frac{L(R - \frac{LB}{\mu})}{\mu E^2}
\end{equation}

In conclusion, the fairness can be estimated by
\begin{align}
    \faa_{avg}(P) &\geq \mathcal E_{E}(\bar\theta) - \mathcal E_{1}(\bar\theta)\\
    &\geq \Big(\frac{E-1}{E} - \frac{L}{\mu E^2}\Big) R
    - \Big(1 +\frac{L(E-1)}{\mu E} - \frac{L^2}{\mu^2 E}\Big) B
    - \frac{2L}{\mu E}\sqrt{B(R - \frac{L}{\mu} B)}
\end{align}
\end{proof}

\subsection{Proof of Divergence Reduction}
\label{appendix:B3}
Here we prove the claim that the assumption $\mathcal L_E(\theta_{e}^*) - \mathcal L_E(\theta_E^*)\geq R$ is implied by a lower bound of the H-divergence \citep{zhao2022comparing}.
\begin{equation}
     D_{H}(\mathcal P_{e}, \mathcal P_E) \geq \frac{LR}{4\mu}
\end{equation}
\begin{proof}
Note that
\begin{align}
    D_H(\mathcal P_e, \mathcal P_E) &= \frac{1}{2} \min_{\theta} \Big(\mathcal L_{e}(\theta) +\mathcal L_E(\theta)\Big) + \frac{1}{2} \Big(\mathcal L_{e}(\theta_e^*) + \mathcal L_{E}(\theta_E^*)\Big)\\
    &\leq \frac{1}{2} \Big(\mathcal L_e(\frac{\theta_e^*+ \theta_E^*}{2}) + \mathcal L_E(\frac{\theta_e^*+ \theta_E^*}{2})\Big) - \frac{1}{2} \Big(\mathcal L_{e}(\theta_e^*) + \mathcal L_{E}(\theta_E^*)\Big)\\
    &\leq \frac{1}{2} \times (\frac{1}{2} L \|\frac{\theta_E^* - \theta_e^*}{2}\|_2^2 \times 2)\\
    &= \frac{1}{8}L\|\theta_E^* - \theta_e^*\|_2^2
\end{align}

Therefore,
\begin{align}
    \mathcal L_{E}(\theta_e^*) - \mathcal L_E(\theta_E^*)&\geq \frac{\mu \|\theta_E^* - \theta_e^*\|_2^2}{2}\\
    &\geq \frac{\mu}{2}\frac{8D_H(\mathcal P_e, \mathcal P_E)}{L} = R.
\end{align}
\end{proof}







\section{Broader Impact}\label{app:impact}

This paper presents a novel definition of fairness via agent-level awareness for federated learning, which considers the heterogeneity of local data distributions among agents. We develop \fairname as a fairness metric for Federated learning and design \name algorithm to improve the corresponding fairness. We believe that \fairname can benefit the ML community as a standard measurement of fairness for FL based on our theoretical analyses and empirical results.

A possible negative societal impact may come from the misunderstanding of our work. For example, low \fairname does not necessarily mean low loss or high accuracy. Additional utility evaluation metrics are required to evaluate the overall performance of different federated learning algorithms.
We have tried our best to define our goal and metrics clearly in \cref{sec:method}; and state all assumptions for our theorems accurately in \cref{sec:theoretical} to avoid potential misuse of our framework.

\end{document}